\documentclass{nwkltrs}

\sidecaptionvpos{figure}{c}

\begin{document}

\title{Artificially intelligent agents in the social and behavioral sciences: A history and outlook}
\headertitle{AI agents in the social and behavioral sciences}

\author[1,2]{Petter Holme}
\author[3,4]{Milena Tsvetkova}

\affil[1]{Department of Computer Science, Aalto University, Espoo, Finland}
\affil[2]{Center for Computational Social Science, Kobe University, Kobe, Japan}
\affil[3]{Department of Methodology, London School of Economics and Political Science, London, United Kingdom}
\affil[4]{Complexity Science Hub, Vienna, Austria}

\twocolumn[%
\begin{@twocolumnfalse}
  \vspace{-15mm}
\maketitle
\begin{abstract}
\noindent We review the historical development and current trends of artificially intelligent agents (agentic AI) in the social and behavioral sciences---from the first programmable computers, and social simulations soon thereafter, to today's experiments with large language models. This overview emphasizes the role of AI in the scientific process and the changes brought about, both through technological advancements and the broader evolution of science from around 1950 to the present. Some of the specific points we cover include: the challenges of presenting the first social simulation studies to a world unaware of computers, the rise of social systems science, intelligent game theoretic agents, the age of big data and the epistemic upheaval in its wake, and the current enthusiasm around applications of generative AI. A pervasive theme is how deeply entwined we are with the technologies we use to understand ourselves.
\end{abstract}
\vspace{5mm}
\end{@twocolumnfalse}]
\thispagestyle{empty}

\section{Introduction}

\begin{SCfigure*}[0.4]
\includegraphics[width=1.46\linewidth]{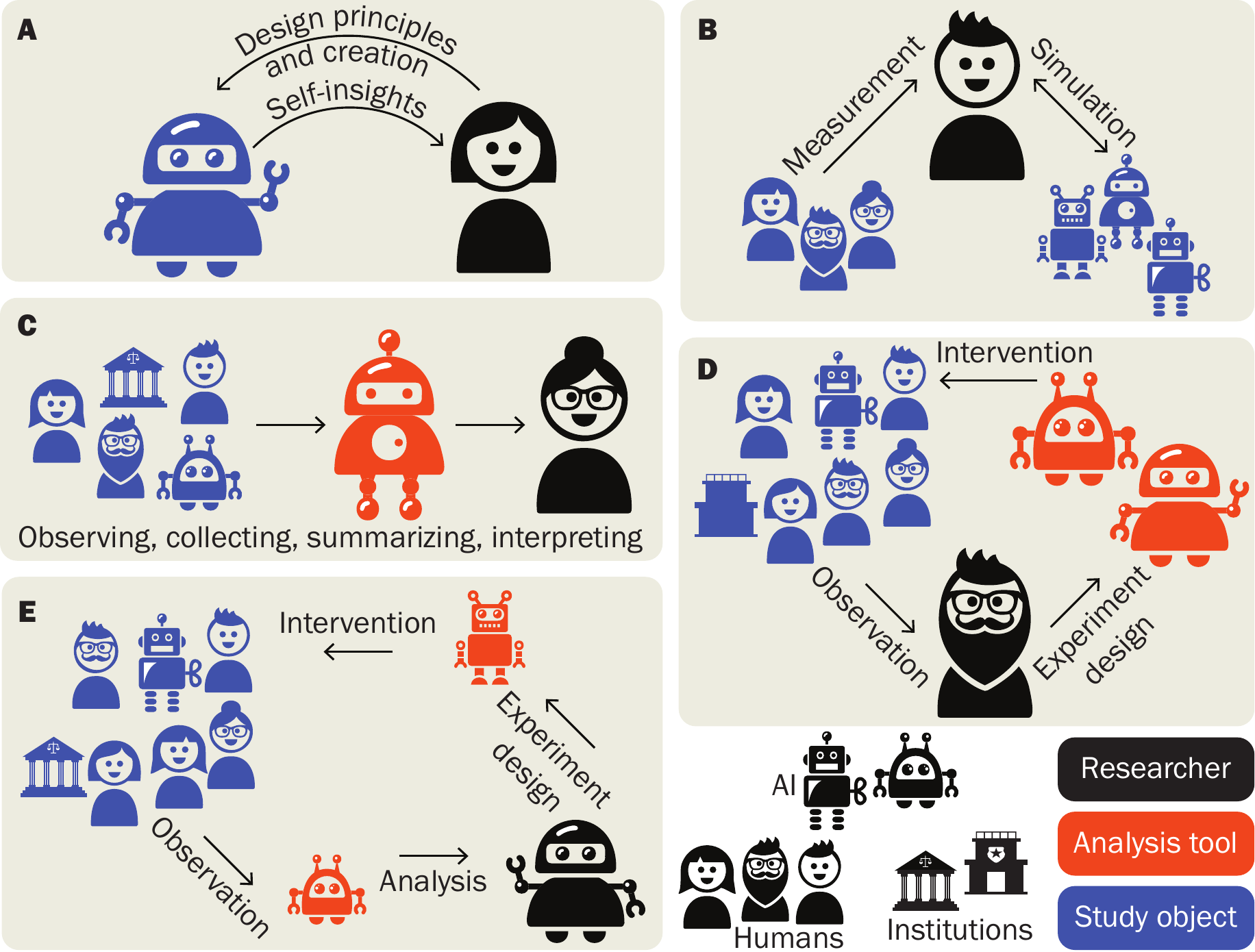}
\caption{\textbf{Illustration of knowledge-building information flows contributing to the behavioral and social sciences involving humans and AI.} Panel A illustrates how AI has always been modeled on human intelligence. This has also shaped our understanding of ourselves. B shows the typical simulation setup in which AI agents mimic a social situation. In C, AI aids in collecting data. Panel D shows experiments in which AI agents interact with humans. Panel E illustrates an AI scientist building knowledge without human intervention. Even though AI-generated science exists today, this particular scenario might still be fictitious, especially given the ethical implications of AI probing human life through interventions~\cite{science2025unethical}.}
\label{fig:ill}
\end{SCfigure*}

For millennia, humanity has cherished a dream of creating machines that can think and behave like ourselves~\cite{mayor}. The closer we get to realizing that goal, the more familiar the image in the mirror appears, and the stronger we believe that understanding that image is really understanding ourselves. The idea of using artificial human replicas in the social and behavioral sciences has seemed self-evident to scientists throughout the ages---every major related technological breakthrough has soon been followed by applications to research on humans. At the time of writing, advances in artificial intelligence, with buzzwords like generative AI (GenAI) and large language models (LLMs), are frequent headline news even outside of science. No wonder this has created yet a wave of research aiming to understand social and behavioral phenomena through new technologies~\cite{evans,christakis_et_al,bail,ziems2023can}. However, this time might be different. For the first time, we have access to ``strong AI'' agents~\cite{searle} that seemingly can take almost any of our roles in experiments and observations (Fig.~\ref{fig:ill}). Then again, this new technology changes our society and behavior in a feedback loop that obscures the road ahead. This paper attempts to cover the developments over the last seventy-five years, leading up to the present use of AI as a research tool. We will emphasize that there has been a bidirectional flow of ideas between AI and the social/behavioral sciences ever since the inception of the former, and try to rectify the picture of AI as a tool arriving to the social/behavioral sciences from afar~\cite{gao2024}.

Telling the story of AI and the social/behavioral sciences is challenging not only because of the entwined flows of ideas and influence, but also because of the many roles AI can play within them. First, as alluded to above, one could see AI as a building block of the scientific method, as an analytical tool, or as a digital twin technology (i.e., a black-box rendition of people or institutions). Second, one could approach AI agents as social actors and try to understand, for example, how behaviorism (which influenced Turing's test~\cite{turing,Boden2006}) transpires in LLMs and GenAI. Third, one could see the emergence of AI as a social phenomenon, in which the imaginations of investors, futurists, and scientists have aligned to bring us to where we are today~\cite{vc_ai,nilsson}. Finally, one could view AI as a social force that changes both interactions between people and people themselves~\cite{yakura_language,farrell2025}. The availability of new research tools aside, understanding this emergent human-machine society is the most pressing motivation for experimenting with agentic AI at the time of writing~\cite{machine_behavior,PEDRESCHI2025104244,tsvetkova_human_machine_networks,tsvetkova2024new}. In Fig.~\ref{fig:ill}, we sketch some approaches in the literature.

The main thesis of this paper is that AI's contribution to the social and behavioral sciences has been highly complex, driven by a multitude of processes, two of which stand out. First, any technical breakthrough is typically rapidly adopted by open-minded social and behavioral scientists, who use it to design new types of investigations. Second, these lines of research branch off to follow a slower-paced evolution centered around the scientific content. For this reason, it is hard to present the historical developments as a timeline. Rather, several computational or AI-infused streams run in parallel. We roughly cover these by their start dates and note that they are all still active research areas to the present. 

Before proceeding to describe the history and present of AI in the social and behavioral sciences, let us say a few words about what this paper is and is not. First, not all the historical approaches we present are based on intelligent agents; however, they all help to set the direction, scope, expectations, and methodologies of agentic-AI science as it stands today. Second, we will not give a comprehensive review of the recent literature as other papers done so~\cite{gao2024}, but we will sketch the current state and trends. Finally, even though there is an overlap between the behavioral and cognitive sciences, we will not cover that interface very deeply---that is indeed a rich and fascinating topic in its own right~\cite{minsky,miller_revolution,minsky1982why,Boden2006}, but more philosophical compared to our paper.

\section{Simulations and early agent-based models}

When the first programmable computer, the ENIAC, began operating in the wake of World War II, it was naturally used for military purposes~\cite{haigh2014alamos}. It is remarkable that only about five years later, the first social-science Ph.D. thesis based on computer simulations was presented~\cite{hagerstrand1953}. This speed and ease of adoption of simulation can be explained in part by the fact that many of the ideas behind electronic computers were widely circulated at the time. For example, the idea that social agents could best be modeled as part random and part purpose-driven was already established~\cite{rashevsky}. How to analyze the output from computer simulations could, of course, be imported from experimental and observational science. Computer simulation manifests a mechanistic mode of scientific explanation---the simulation connects microscopic observations to produce verifiable macroscopic patterns. The logic of this form of explanation existed long before the advent of computer simulations---see, e.g., Ragnar Frisch's microscopic models of business cycles from the 1930s~\cite{frisch}. The transition was indeed so smooth that there were papers based on stochastic simulations using manual random number generation, such as Helen Abbey's 1952 study of the Reed-Frost epidemic model~\cite{abbey1952}.

Early computer-aided studies in the behavioral and social sciences, of course, faced some problems that we do not have today---this was a time when computers were something people might have read about in a newspaper but usually never even seen in their everyday lives. In 1966, authors could feel compelled to explain computers as~\cite{gilbert1966}:
\begin{quote}
\dots a clerk with several virtues, but one which is no better than its instructions. To appreciate the way in which this clerk operates, one should realize that a computer is, in principle, nothing more than a desk calculator and a notepad.
\end{quote}

With describing these transistorized clerks out of the way, the next challenge was to convince the readers of their methodological merits. In this aspect, the papers have a much more modern feel. We can notice five such themes:

\paragraph{Nothing new}
Several early papers point out that simulating humans did not start with digital computers. Harry Harman wrote in a military white paper in 1961 that~\cite{harman1961}:
\begin{quote}
Simulation may be traced back to the beginning of time---be it the make-believe world of the child at play, or the adult make-believe world of the stage.
\end{quote}
In his \textit{The Sciences of the Artificial}, Herbert Simon makes the same point by referring to an analog simulator of Keynesian economics (the \textsc{Moniac})~\cite{simon1969}.

\paragraph{To facilitate systems thinking}
As we will discuss further below in the context of systems science, one radical insight in 20th-century science was that when a system contains feedback loops, so things affect themselves, one cannot discuss causality in usual terms~\cite{cybernetic_moment}. Mathematical models get hard to solve, but, as some early papers point out, could still be analyzed by computer simulations~\cite{gullahorn1965,zelditch_evan}.

\paragraph{To do experiments otherwise impossible}
This was perhaps the most common motivation and one that has fallen somewhat out of fashion. Quoting the psychologist Kenneth Colby (1967)~\cite{colby1963}:
\begin{quote}
Before the computer program we had no satisfactory approach to huge, complex, ill-defined systems difficult to grapple with, not only because of their multivariate size but also because of a property of elusiveness which in psychology is mainly a function of vagueness in that the limits of inclusiveness of conventional terms are unclear.
\end{quote}
The ``complex, ill-defined system'' Colby had in mind was nothing less than a ``neurotic patient.'' For another example, de Sola Pool and Kessler use this motivation in their essay ``The Kaiser, the Tsar and the Computer'' to present their simulation-based decision support system for international political conflicts, Crisiscom~\cite{desola1965}.

\paragraph{Proof-of-concept models}

Another motivation for computer simulations in the 1950s and '60s was that they could help validate qualitative theorizing. What good is a theory if it cannot even reproduce qualitative results? AI pioneer Edward Feigenbaum wrote in his 1963 essay ``Computer simulations of human behavior''~\cite{feigen}:
\begin{quote}
One of the advantages of computer simulation is this one, of guaranteeing sufficiency and completeness. The computer simulation model will not operate if you forget anything. If you fail to take into account some necessary mechanism that, in the verbal description of a theory, you might readily pass over, the computer simulation will not run. Thus, one is forced to focus attention on all of the mechanisms---those that are not well understood as well as those that are well understood---and all of the subtle effects and interactions which are implicit in one's model. In a sense, this business of accounting for all of the information processing is a very powerful mental discipline.
\end{quote}

\paragraph{For forecasting and scenario testing}

Forecasting and scenario testing are, of course, areas where simulations could inform decision-makers. The above-mentioned crisis simulator by de Sola Pool is a good example and may be the reason political science was very early to jump on the simulation bandwagon~\cite{desola1965}. Another example by the same authors was a simulation program for American presidential elections~\cite{pool1965}, which, at the time, played a similar role to that of election predictors and prediction markets today~\cite{wolfers2004prediction}.

\section{Systems and information}

Another side of the intellectual and scientific trends that led to the development of the electronic computer was a focus on systems---collections of units that cannot be understood by reducing them to simple causal relations~\cite{cybernetic_moment,heims,ashby,wiener_human_use}. One academic movement based on this tenet, predating artificial intelligence, was \textit{cybernetics}. Cybernetics encompassed topics such as control, communication, information transfer, and an aspiration to not only transcend scientific disciplines but also break the barriers between the applied and theoretical sciences.

Cybernetics influenced the social and behavioral sciences in several waves. The first wave was centered around ten so-called Macy conferences~\cite{macy_conference} in 1946--53. Some of these meetings had an explicit social theme. For example, the second conference gathered sociologists Paul Lazarsfeld, Talcott Parsons, and Robert Merton, psychologist Kurt Lewin, and anthropologists Margaret Mead and Gregory Bateson, to discuss teleological mechanisms in society. Here, ``teleological'' referred to Rosenblueth, Wiener, and Bigelow's then-recent essay ``Behavior, purpose and teleology''~\cite{rwb}, which argued for cybernetics as an alternative to behaviorism.\footnote{The cyberneticians saw behaviorism as overly simplistic in its inability to handle causal feedback loops.} Even if the first generation of cyberneticians didn't directly inspire computational approaches to the social sciences, their systems thinking had an impact on, e.g., Parsons'~\cite{parsons} and Merton's~\cite{merton} work from around this time.

It is hard to pinpoint the first systems-oriented computational framework, but the most successful of the early ones was arguably Jay Forrester's \textit{systems dynamics} from the late 1950s~\cite{forrester_industrial}. Inspired by the management of military operations, Forrester proposed coupled differential equations with delays as a universal systems-simulation paradigm and went on to make an unreplicated political impact. In the early 1970s, the environmentalist movement had its first peak, the public belief in computer technology was strong, and systems thinking was making an academic impact~\cite{odum}. All this set the stage for the report \textit{Limits to Growth}~\cite{meadows1972} that presented system dynamics simulations of the global population, food supply, and environmental pollution to argue that the economic growth paradigm is doomed to failure in our world of limited resources~\cite{hegemony}. \textit{Limits to Growth} had a tremendous political impact, and one can argue that today's degrowth movement~\cite{degrowth}, the UN's Sustainable Development Goals, and academic research in Earth system analysis can all be traced back to it~\cite{hegemony}.

\begin{SCfigure*}[0.5]
\includegraphics[width=1.667\linewidth]{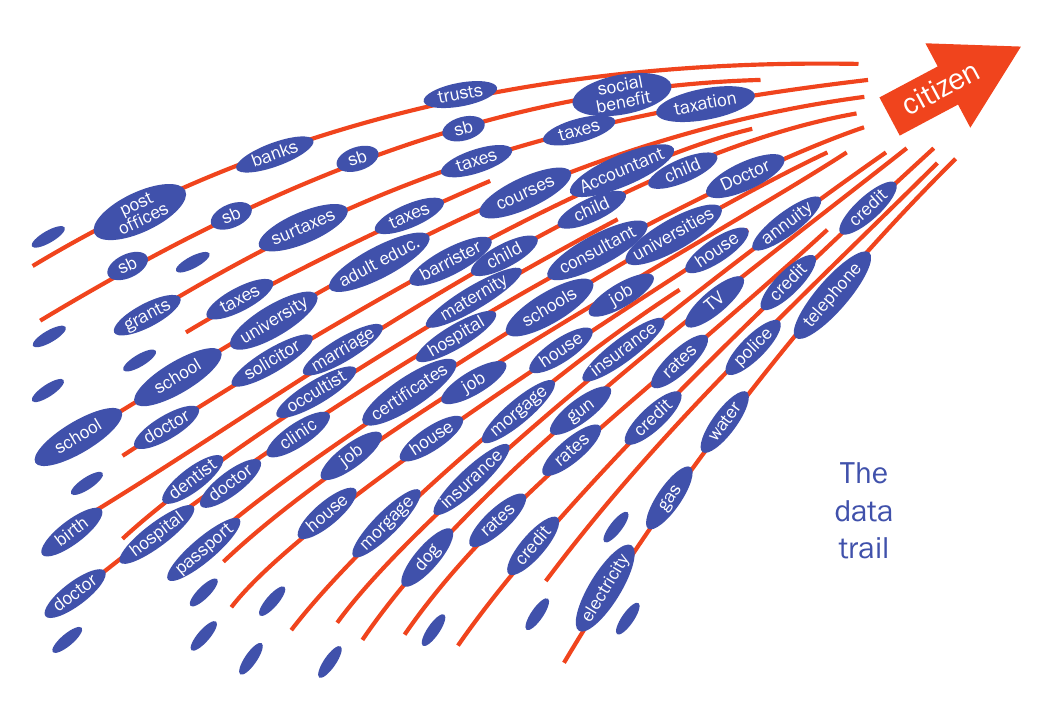}
\caption{\textbf{An illustration of trace-data to support governance} adapted from Stafford Beer's 1975 \textit{A Platform for Change}~\cite{beer}. Even though Beer envisioned a planning economy capable of catering to the needs of the individual via data mining and cybernetic control~\cite{medina}, this figure inadvertently seems to anticipate today's information economy and digital surveillance~\cite{lyon,kalluri2025computer}.}
\label{fig:beer}
\end{SCfigure*}

In the 1960s, cybernetics had a computerized revival, especially in Britain~\cite{pickering}. Notably, Stafford Beer and colleagues presented cybernetic support tools for governance that foreboded today's information economy and computational social science, see Fig.~\ref{fig:beer}. Beer also almost got the chance to put his theories to work in the socialist regime of Chile's Salvador Allende, where he led the development of the governance support system Cybersyn~\cite{medina}.\footnote{Allende's regime came to a dramatic end with the 1973 \textit{coup d'\'etat}, but interestingly, the new regime under General Augusto Pinochet also played a minor part in our story. The military leadership consulted the Nobel laureate and proponent of laissez-faire economics, F. A. Hayek, to rebuild the Chilean economy---the same Hayek who helped lay the foundations for the currently prevailing connectionist approach to AI (see below, and Ref.~\cite{hayek}).}

\begin{SCfigure*}[0.7]
\includegraphics[width=1.3\linewidth]{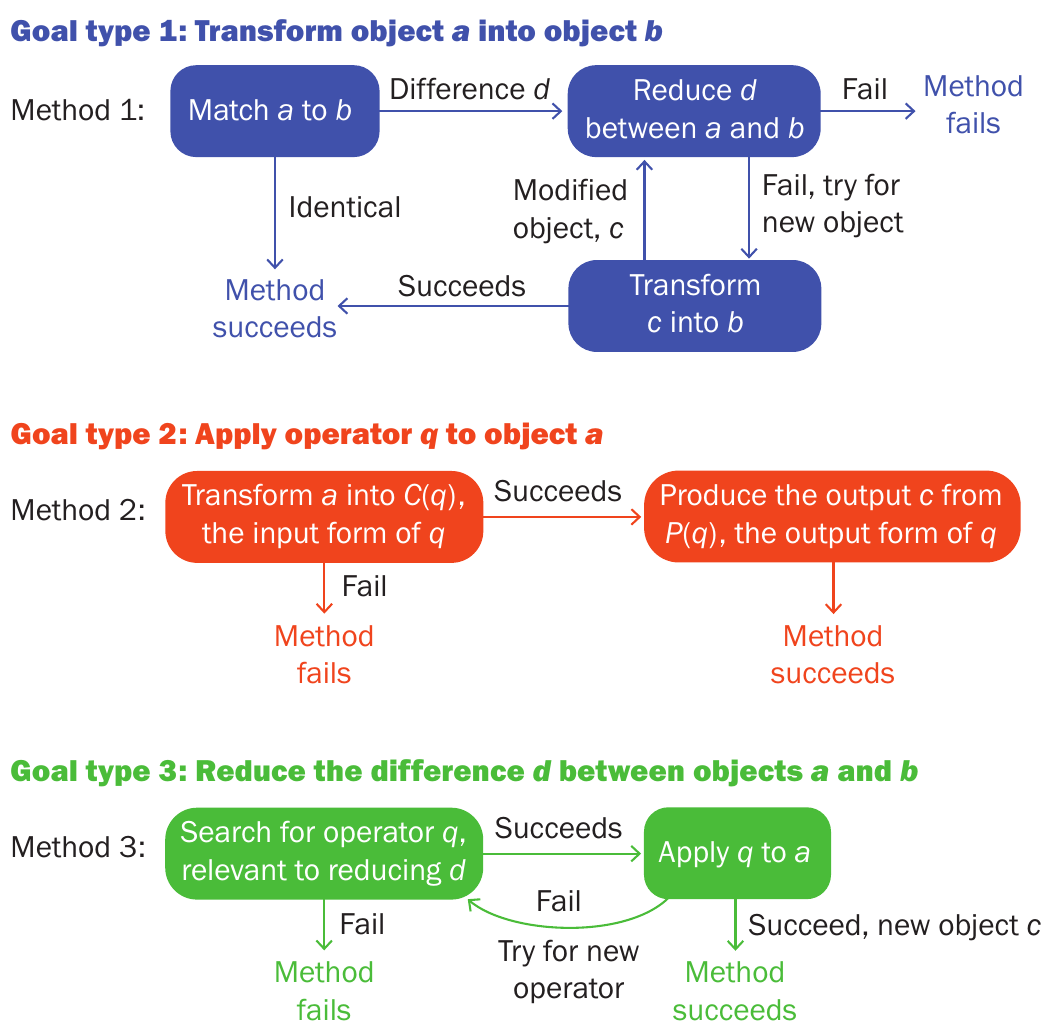}
\caption{\textbf{Part of the flow chart of Newell and Simon's General Problem Solver}~\cite{newell1961}---the part related to ``means-end analysis.'' Without going into detail about how the constituents are manifested in the program, we can still appreciate how different Newell and Simon's approach is from today's machine-learning-based AI. Newell and Simon explicitly stated that Turing was not an inspiration for their AI~\cite{mccorduck}, which seems evident from the General Problem Solver.}
\label{fig:gps}
\end{SCfigure*}

\section{Pioneers of artificial intelligence}

Even though many foundational works were published earlier---like Turing's ``Computing machinery and intelligence''~\cite{turing} and the neuronal model of McCulloch and Pitts~\cite{mcculloch_pitts}---the academic field of artificial intelligence usually sets its birth date to 1956. This was the year when John McCarthy and Marvin Minsky organized a research program titled ``Dartmouth Summer Research Project on Artificial Intelligence.'' The term artificial intelligence was allegedly a political choice to stand out from neighboring fields~\cite{mccorduck}.\footnote{Indeed, several of the fields and subfields mentioned in this article are primarily social rather than topical constructs.}

Throughout its history, AI has included a wide variety of approaches to machine intelligence. It has also had many ups and downs, and the term's connotations have fluctuated in concert with the paradigm of the day. From the outset, some researchers, such as Allen Newell and Herbert Simon, approached AI by reverse-engineering higher-order cognitive abilities such as reasoning and problem-solving; others, such as Walter McCulloch and Frank Rosenblatt, focused on the computational capabilities of interconnected nerve cells. These roughly correspond to the symbolic and connectionist approaches to AI~\cite{mitchell_artificial_2019,aima}.

\subsection{Simon \textit{et al.}'s symbolic reasoning AI}

Herbert Simon's career started with the study of administrative behavior and organizational decision-making~\cite{simon_admin}. Although Simon's output was extremely diverse, one can often trace his ideas back to organizational decision processes of one sort or another~\cite{simon_my_life}. At the aforementioned Dartmouth conference, Simon and his long-standing collaborator Allen Newell presented their theorem-proving program ``The Logic Theorist'' that was capable of solving logical propositions and is sometimes regarded as the first AI program~\cite{mccorduck}. Continuing this line, Newell and Simon built the General Problem Solver (GPS)---a program simulating human reasoning in general terms~\cite{newell1961} (see Fig.~\ref{fig:gps}).

Simon was later awarded the Nobel Economics Prize primarily for his work on bounded rationality~\cite{simon_bounded}. However, Simon's social science most pertinent to this paper was presented in \textit{The Sciences of the Artificial}~\cite{simon1969}, which was foundational to both design theory and the study of complex social systems. One often misunderstood aspect of this work today is that Simon called for an investigation of the design process of artificial objects and their \textit{de facto} usage, not just their functionality~\cite{simon1969}:
\begin{quote}
The thesis is that certain phenomena are ``artificial'' in a very specific sense: they are as they are only because of a system’s being molded, by goals or purposes, to the environment in which it lives. If natural phenomena have an air of ``necessity'' about them in their subservience to natural law, artificial phenomena have an air of ``contingency'' in their malleability by environment.
\end{quote}

While the AI of Simon \textit{et al.}\ may seem a bit outdated today, it did have an immense impact, much in the form of ``expert systems''~\cite{durkin}. Expert systems constitute perhaps the first commercially successful type of AI application, with users primarily in corporate and institutional management. Their advances in knowledge representations continue to influence fields such as knowledge management and organizational learning to this day~\cite{nonaka2006organizational}.

In a prescient work~\cite{plural_soar}, Kathleen Carley and colleagues attempted to connect Allen Newell's ``Soar'' architecture for a general-purpose AI (or \textit{cognitive architecture} as they termed it)~\cite{newell_unified} into a model of organizational decision making. Carley \textit{\textit{et al.}}\ considered agents who need to find components in a warehouse to complete incoming orders and require both coordination and communication to achieve this. See Table~\ref{tab:carley} for a list of features needed to replicate a human in a social simulation according to Ref.~\cite{plural_soar} (a topic Carley and Newell expanded on in Ref.~\cite{Carley01121994}).

\begin{table}
\caption{\textbf{Required capabilities of social agents} according to Carley and Newell~\cite{plural_soar}. The authors admit that they have managed to achieve all except the ones pertaining to ``social analysis.''}
\vspace{3mm}
\begin{tabular}{l}
\hline
\textit{Perception and Action} \\
\hspace{6mm} Perceives the environment \\
\hspace{6mm} Physically manipulates objects \\
\hspace{6mm} Moves self to different locations \\
\textit{Memory} \\
\hspace{6mm} Location \\
\hspace{6mm} People \\
\hspace{6mm} Task \\
\textit{Instruction} \\
\hspace{6mm} Can be incomplete \\
\textit{Task Analysis} \\
\hspace{6mm} Decomposes task \\
\hspace{6mm} Coordinates subtasks for self to do \\
\textit{Communication Skills} \\
\hspace{6mm} Asks questions / Provides answers \\
\hspace{6mm} Gives commands / Receives commands \\
\hspace{6mm} Talks to a single individual / Talks to a group \\
\textit{Social Analysis} \\
\hspace{6mm} Models of other agents \\
\hspace{6mm} Model of organization\\
\hline
\end{tabular}
\label{tab:carley}
\end{table}

\subsection{The early connectionists}

The other major branch of early artificial intelligence, \textit{connectionism}, was the antecedent of many of the most hyped-up technologies of today~\cite{aima}. This is the tradition of building machine intelligence on networks of model neurons. The starting point of connectionism was the \textit{perceptron}---a neuronal model invented by McCulloch and Pitts in 1943~\cite{mcculloch_pitts}. Fifteen years later, Frank Rosenblatt implemented a network of perceptrons with custom-made hardware~\cite{rosenblatt_perceptron}. Rosenblatt's paper is an interesting read, going far beyond technicalities and crediting, among others, the above-mentioned F. A. Hayek for his philosophically trailblazing \textit{The Sensory Order}~\cite{hayek}. After their initial success, artificial neural networks entered a hiatus that would last throughout the 1970s~\cite{feldman_ballard}.

\section{Neo-cybernetics}

In this section, we will cover some loosely connected streams of ideas that link the original cybernetics movement to the AI-infused, human-centric science of today. It contains more engineering-minded research, aimed at improving the world, compared to the following section.

\subsection{Embodied cognition and autopoiesis}

The approaches to artificial intelligence and modeling of humans described above (except, perhaps, the perceptron) share the fact that they do not depend on a physical body. They are consistent with the Cartesian dictum that the mind and the body can be considered separate, with the mind operating solely as an information-processing device. This orthodoxy has been challenged several times: Merleau-Ponty's \textit{phenomenology of perception} in the 1940s~\cite{merleau_ponty}, Gibson's \textit{ecological psychology} and his concept of \textit{affordance} of the 1960s~\cite{gibson}, and by the early 1990s, \textit{embodied cognition}~\cite{varela1991embodied,wilson}. The latter also became influential in robotics, where, logically, artificial intelligence should be localized to the robot itself, in its environment~\cite{froese2009,PfeiferScheier1999}.

From the same spirit of skepticism as above---maybe things that were considered separate actually are sufficiently connected to be regarded as one---cybernetics had another energy injection in the early 1970s with \textit{second-order cybernetics}, making the point that an observer cannot be fully detached from the observed object or system~\cite{mead}. It is not a coincidence that this line of thinking coalesced with the ever more inclusive system-simulation models (e.g., \textit{The Limits to Growth})~\cite{odum,meadows1972,roberts}. Furthermore, the inevitable consequence of allowing wider and looser boundaries is that one's study objects will have more capabilities. Against this backdrop, Humberto Maturana and Francisco Varela arrived at their concept of \textit{autopoiesis} in 1972~\cite{autopoiesis}. This refers to the ability of living systems to reproduce themselves by organizing, assembling, and, to some extent, producing their constituents. Autopoiesis has had a broad influence---from manufacturing engineering, via philosophy of mind, to sociology. In the latter, thinkers like Niklas Luhmann~\cite{luhmann1986autopoiesis} and Edgar Morin~\cite{morin} explored the autopoietic nature of social systems.

\subsection{Human-machine interaction}

 Although it is primarily concerned with advancing technology rather than our understanding of society and human behavior, the field of human-computer interaction (HCI) has much overlap with the themes in this paper. One influential opinion/framework paper in this area was Clifford Nass's 1994 ``Computers are social actors''~\cite{casa}, arguing that since people naturally anthropomorphize computers, it makes sense to approach HCI with methods from the behavioral and social sciences. Influenced by embodied cognition, this research expanded into topics such as the integration of socially aware agents into human populations~\cite{dautenhahn,sebo2020robots}. Furthermore, it approximated today's discussion of AI in society, and even made some points that have since been lost. For example, this research stressed the importance of embodied perspectives and narrative understandings of courses of events if one seeks to make AI capable of replacing humans.

\begin{figure*}
\centering\includegraphics[width=0.9\linewidth]{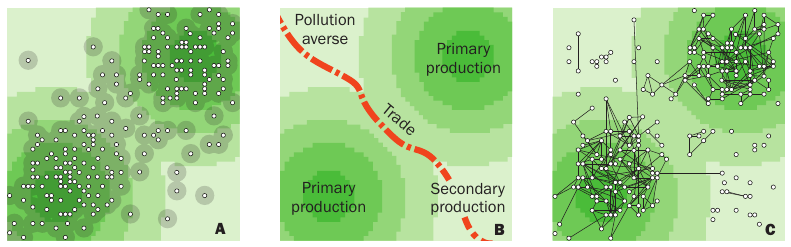}
\caption{\textbf{An illustration of the Sugarscape model} by Epstein and Axtell~\cite{sugarscape}---an archetypal agent-based model. The simulation is based on agents spread out over a grid, and an unevenly distributed natural resource---the ``sugar.'' A darker shade of the background color indicates a higher sugar concentration. In principle, one could use any grid geometry and sugar distribution, but Sugarscape has become synonymous with the shown bimodal distribution on a 50 $\times$ 50 grid. Agents move to optimize a utility function within their ``field of vision''---the gray surroundings of the white dots in panel A. If direct sugar consumption is all the agents care about, their population distribution will come to resemble the sugar distribution (panel A). By adding abilities to the agents and terms to their utility functions, the Sugarscape simulation can recreate many spatial features and functions of a society and economy, including the emergence of ethnicities, borders, border trade, a segregation of people by their aversion toward pollution, etc.\ (panel B). It is also possible to extend the simulation with non-local relationships between agents, to make Sugarscape the basis of a social network simulation.}
\label{fig:sugarscape}
\end{figure*}

\begin{figure*}
\centering\includegraphics[width=\linewidth]{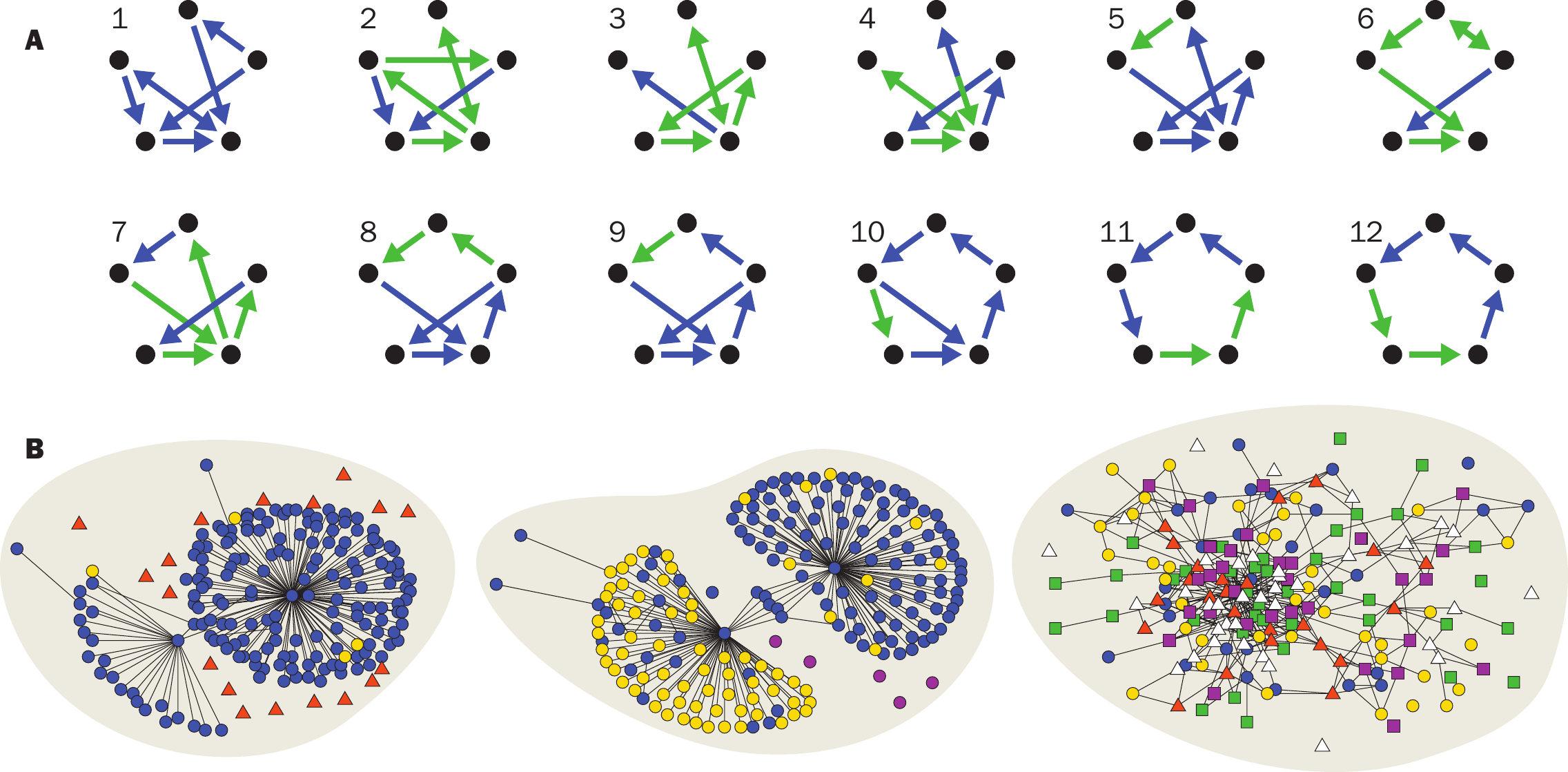}
\caption{\textbf{Example output from network models of agents trying to optimize their network positions}, adapted from Refs.~\cite{bala_goyal} and \cite{diplo}. Panel A illustrates an economics-style model where agents optimize a utility function that represents a trade-off between the number of agents whose information can reach them (directly or indirectly through paths of directed links) and the in-degree (i.e., the number of direct-link neighbors)~\cite{bala_goyal}. The figure shows agents updating their links until a Nash equilibrium (which is an absorbing state) is reached (at time step 11). The updating proceeds with agents either configuring their network to what would have been optimal at the previous time step (green arrows) or, with some probability, doing nothing (blue arrows). Panel B presents snapshots of a model with agents again attempting to optimize conflicting objectives: namely, simultaneously maximizing their closeness centrality (a proxy for power) and minimizing their degree (the cost of communication)~\cite{ghoshal_holme}. The agents can reconfigure their positions locally (within their second neighborhoods) and use reinforcement learning based on their own and their neighbors' performance to decide their updating strategy. The updating strategies are codified as a priority list of actions to take (e.g., deleting the link to the lowest-degree neighbor). The symbols represent actions of highest priority for the respective agent.}
\label{fig:nwks}
\end{figure*}

\section{Complexity science and game theory}

Around the mid-1980s, another research field, \textit{complexity science}, appeared on the scene. Although it thematically overlaps with systems science, cybernetics, etc., complexity science had, at the outset, other sources of inspiration: chaos theory, fractals, cellular automata, and a shared sense of the importance of interdisciplinary research~\cite{waldrop,holme_arxiv}. There were (and still are) deeper differences between systems and complexity science. Systems science, on the one hand, primarily focuses on closed systems with irreducibly complicated interactions (e.g., autopoiesis in organizational engineering~\cite{Schatten2014}). Complexity science, on the other hand, stresses complex patterns emerging~\cite{holland} from simple rules (like, say, Conway's ``game of life'' being capable of universal computation~\cite{game_of_life}). The way emergence has been conceptualized in complexity science since the mid-1990s has deep connections to the social sciences~\cite{new_social_science}, being akin to Thomas Schelling's and James Coleman's micro-to-macro transitions~\cite{coleman1986,schelling_micromotives_1978,scott_page}, and F. A. Hayek's ``spontaneous order''~\cite{hayek1969studies}.

\subsection{Agent-based models}

While complexity science has always had one foot (out of more than two) in the social sciences\footnote{For example, Brian Arthur's complexity economics~\cite{arthur} or urban scaling theory~\cite{bett}.}, its relevance for our story mostly comes from agent-based models (ABMs). These were not invented in complexity science---for example, some of the social simulation models mentioned above are ostensibly agent-based---but complexity science was the first discipline to treat agent-based modeling as a comprehensive framework connecting micro- and macro-level phenomena.

Traditionally, the behavioral responses of agents have been encoded by simple decision rules~\cite{gilbert_troitzsch,gilbert_conte}. Although always considered unorthodox, ABMs have found their way into relatively conventional social science~\cite{macy2002factors,sawyer}. A notable example is their use in analytical sociology~\cite{hedstrom,hedstrom_abm} and Epstein and Axtell's more interdisciplinary \textit{Sugarscape} simulations~\cite{sugarscape} (see Fig.~\ref{fig:sugarscape}). Economics has also seen a use of ABMs~\cite{farmer_abm,lettau1997explaining}, perhaps most frequently in the computational finance literature~\cite{lebaron2000agent}. These studies include the use of reinforcement learning neural-network agents~\cite{beltratti1996neural}. Finally, management science, which was keen to import complexity science concepts such as the ``edge of chaos''~\cite{surfing}, ventured into the realm of ABMs as well~\cite{abm_business}, especially to study innovation~\cite{levinthal,FRENKEN2000257} with the so-called NK model of evolution~\cite{nk_model}.

\subsection{Game-theoretic agents}

The use of ABMs in political science~\cite{axelrod1984}, economics~\cite{abm_economics}, and, to some extent, anthropology~\cite{anasazi} usually rests on a game-theoretical foundation~\cite{textbook_in_economic_game_theory}, assuming people make decisions by maximizing utility functions. When different time horizons clash in the agents' utility optimization, we arrive at a \textit{social dilemma}---a situation where the agents are faced with a choice to behave selfishly (typically leading to higher profits in the short term) or prosocially (engaging in cooperative efforts to build society). The study of human cooperation from this perspective has evolved in peculiar ways, with different sub-communities within the field splitting and later merging again. In the 1980s, the inspiration came from population biology~\cite{maynard_smith}, with the sedentary transition of human prehistory as the explanandum~\cite{boehm}. In the early 1990s, the mathematics of complexity emerging from simplicity was a major driving force in research~\cite{nowak_may}. In parallel to these developments, social dilemmas have been studied in economics in several waves. Notably, starting with Fehr \textit{et al.}'s work on punishment and public goods~\cite{fehr2000cooperation}, the early 2000s saw increased interest in behavioral economics experiments on free-riding and the over-exploitation of shared resources. Some of these experiments have mixed human and bot populations, like Fig.~\ref{fig:ill}C, usually with fairly simple rule-following bots~\cite{march2021strategic}. These usually present ways bots can improve team performance~\cite{vaccaro2024when}. For example, Shirado and Christakis show that randomly behaving bots can push human participants out of local optima and toward better performance in a color-coordination game~\cite{shirado_christakis}.

\subsection{Artificial life}

Yet another wildly different field that emerged from complexity science and is relevant to us is \textit{artificial life}~\cite{stephen_levy}. This field---with roots in John von Neumann's search for self-replicating cellular automata~\cite{von_neumann_automata}---attempts to create computer programs with properties analogous to those of living organisms. Although artificial life rarely deals with higher-order cognitive processes, the initial hype around a series of conferences organized by Chris Langton at the Santa Fe Institute~\cite{langton} was highly reminiscent of  today's debates about the hypothesized consciousness of LLMs and the purported impending singularity induced by ``artificial general intelligence''~\cite{coming_evolution}:
\begin{quote}
    Within fifty to a hundred years, a new class of organisms is likely to emerge. These organisms will be artificial in the sense that they will originally be designed by humans. However, they will reproduce, and will evolve into something other than their initial form; they will be ``alive'' under any reasonable definition of the word. These organisms will evolve in a fundamentally different manner than contemporary biological organisms, since their reproduction will be under at least partial conscious control, giving it a Lamarckian component. The pace of evolutionary change consequently will be extremely rapid. The advent of artificial life will be the most significant historical event since the emergence of human beings. The impact on humanity and the biosphere could be enormous, larger than the industrial revolution, nuclear weapons, or environmental pollution. We must take steps now to shape the emergence of artificial organisms; they have potential to be either the ugliest terrestrial disaster, or the most beautiful creation of humanity.
\end{quote}
Thirty years later, the field of artificial life still studies similar questions, still primarily \textit{in silico}, but is somewhat more modest in its future predictions~\cite{gershenson_sayama}.

\subsection{Network science and networking agents}

Representing social relations between individuals or institutions as graphs has a long, albeit patchy, history in the social sciences~\cite{freeman2004development}. The basic tenets were outlined in the 1930s by Helen Jennings and Jacob Moreno, and embrace de Saussurean structuralism in its recursiveness: actors' roles are determined by how they are connected to other actors~\cite{jennings_moreno}. This means that if one maps out the network of connections, there should be mathematical ways to characterize the nodes' roles by measuring their positions within the network. Much of social network analysis~\cite{wasserman1994social} and network science~\cite{bara} builds on this idea and thus seeks to develop measures linking positions and roles, structure and function.

The works of Milgram~\cite{travers1969experimental}, Granovetter~\cite{granovetter1973strength}, and others in the 1960s and 70s led to the insight that people are remarkably good at using their social networks~\cite{friends}. Along with the above-mentioned assumption that network positions determine people's opportunities, and the economic dogma that people strive to optimize their prospects, it follows that economic agents should seek to improve their network positions~\cite{sayed,bala_goyal,ghoshal_holme}. Figure~\ref{fig:nwks} shows two models of agents trying to optimize their network positions using different utility functions and updating algorithms. The model in Fig.~\ref{fig:nwks}A adopts an economics-style approach, assuming that agents can calculate their optimal actions. In contrast, the agents of Fig.~\ref{fig:nwks}B learn their networking strategies through reinforcement learning. Note that the network structures of both models fluctuate wildly compared to social networks in the wild, suggesting a model-reality gap to close in future studies.

\section{The age of big data}

Around the first years of the new millennium, the AI winter gave way to a spring and summer that still linger. The one-sentence summary of this development is: By using ever more resources, general methods can vastly outperform earlier, specialized ones~\cite{halevy2009unreasonable}. The period up to today can be divided by the advent of near-human-level chatbots, where this section covers the first (and epistemically more disruptive) part. During this period, there was a concomitant advancement in hardware, data collection, and analysis methods, accompanied by growth in the information economy and an increased acceptance of the deployment of algorithmic decision making in areas such as insurance, corporate hiring, and trading~\cite{mathdestruction,hendershott}. This was fed into a feedback loop that catalyzed societal changes as interactions increasingly occurred online, thereby producing more data to fuel the algorithms. The zeitgeist is well captured by the McKinsey Global Institute's 2011 report touting big data as ``the next frontier for innovation, competition, and productivity''~\cite{manyika2011big}.

The idea of extracting insights from vast streams of data---articulated decades earlier by the likes of Stafford Beer (Fig.~\ref{fig:beer}) and Marshall McLuhan (``information overload = pattern recognition''~\cite{mcluhan})---was finally becoming reality~\cite{sun2017revisiting}. Around 2010, \textit{data science} and \textit{data scientist} became the established terms for this new guild and its members.

\subsection{Advancements in (un)supervised learning}

The big-data era did not happen so much because of breakthroughs in machine learning (ML) algorithms as the ability to scale old ones to orders-of-magnitude larger datasets. The resurgence and mainstream adoption of \textit{deep learning}~\cite{hinton_deep_learning,schmidhuber,lecun2015deep} is probably the best example of this. In 2012, AlexNet, a deep-learning architecture developed by Krizhevsky, Sutskever, and Hinton, won an esteemed computer vision challenge by a legendary margin~\cite{krizhevsky2012imagenet}. AlexNet demonstrated the efficacy of graphics processing units for training deep neural networks, thereby placing renewed emphasis on hardware. Following AlexNet, subsequent key innovations included the development of convolutional neural network architectures such as VGGNet~\cite{simonyan2014very,schmidhuber}, GoogLeNet with inception modules~\cite{szegedy2015going}, and ResNet, which introduced the residual learning framework that enabled unprecedented network depths~\cite{he2016deep}. These developments established deep-learning architectures in computer vision with extensive industrial applications in autonomous vehicles, facial recognition, mass surveillence~\cite{kalluri2025computer}, and image-based diagnostics.

Parallel advancements occurred in natural language processing (NLP), marked notably by the introduction of recurrent neural network architectures like ``long short-term memory'' networks for sequence modeling tasks~\cite{lstm,schmidhuber}. This progress invigorated machine translation, speech recognition, and language modeling~\cite{graves2013speech}. The release of Word2Vec by Mikolov \textit{et al.}~\cite{mikolov2013efficient}, which efficiently generated word embeddings that capture semantic relationships, transformed NLP methodologies and laid the groundwork for later embedding-based models. By 2017, the introduction of transformer models, particularly the seminal ``Attention is all you need'' paper by Vaswani \textit{et al.}~\cite{attention}, took AI closer to being generative in a meaningful way. At the time of writing, almost all new conversational AI platforms build transformer units.

\subsection{The revival of reinforcement learning}

Another ML paradigm that got a revival in the early 2010s was reinforcement learning~\cite{reinforcement}. This concerns, very briefly, how to iteratively learn optimal decision-making. In particular, the highly publicized Go match between the reinforcement-trained deep-learning architecture AlphaGo and Korean Go master Lee Sedol~\cite{silver2017mastering} in 2016 revitalized this then-mature topic.

Since humans and animals also learn to make decisions through successive mistakes, there are deep connections between reinforcement learning and other areas of science. For example, Thorndike noticed in 1911 that cats find a balance between exploitation and exploration in problem-solving experiments~\cite{thorndike1911animal}. This dilemma, whether to stick to the best-known policy (local optimum) or search for even better ones (global optima), is a recurrent theme in fields as diverse as robotics, optimization, economics~\cite{arrow1962economic}, management~\cite{march_exploration_1991}, and finance~\cite{charpentier2023reinforcement}.

\subsection{Computational social science and digital traces}

The initially corporate term ``data science'' was soon adopted by academia, with universities launching degree programs, associations springing up, journals being founded, etc. Some authors see the field of \textit{computational social science} as nascent of this development~\cite{watts2007twenty,lazer2009computational,salganik2017bit}, whereas others point to deeper roots~\cite{holme_lilje,Wallach2018}. Regardless, toward the turn of the millennium, an increasing availability of data more or less directly recording human behavior---\textit{digital trace data}---made the Internet a valuable data source for the social and behavioral sciences~\cite{grimmer_roberts_stewart}, with social media as the chief data source~\cite{kwak2010twitter,sushil_tweet}.

Even if there were precursors over a decade earlier~\cite{early_email}, research with digital traces became popular around the final years of the millennium. Digital trace datasets are distinctive in that they are not gathered primarily for scientific purposes~\cite{watts2007twenty,lazer2009computational}. Moreover, they are often larger than the typical social-science datasets in the earlier literature and thus require alternative analytical methods and different research questions. For these reasons, it is perhaps not so surprising that early papers on digital trace data were typically carried out by people with backgrounds in the natural or formal sciences. As an example of such an early study, Huberman \textit{et al.}~\cite{huberman_www} found a power law in histograms of arrival rates for unique visitors to websites. They went on to explain this observation by a simple random-walk style model, but---as several other early computational social science articles---did not relate it to the bigger social-science context.\footnote{E.g., Arthur's economics of increasing returns~\cite{arthur1989competing} was probably applicable to the online platforms of that era.}

As more social scientists saw the value of digital traces, the early years of soul-searching gave way to a more broadly informed computational social science. By the mid-aughts, the field had matured enough for studies to become more integrated with mainstream social science. Salganik, Dodds, and Watts' study of success and failure on a website for music-sharing is a prime example of this development~\cite{sagalnik_dodds_watts}.

\begin{SCfigure*}[0.6]
\centering\includegraphics[width=1.25\linewidth]{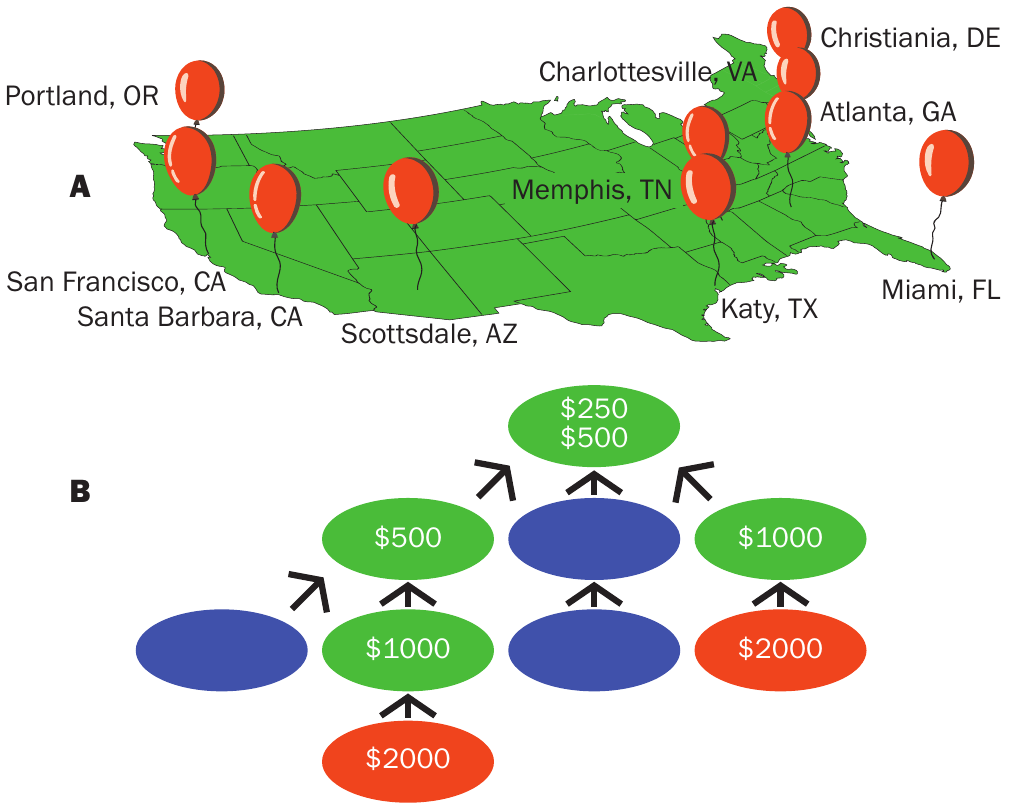}
\caption{\textbf{The DARPA Red Balloon Challenge} was a contest in social mobilization for geo-location to commemorate the Internet's 40th anniversary. On December 5, 2009, ten red balloons were positioned at undisclosed locations across the continental USA (panel A). The first participant to correctly report the location of all was awarded \$40,000. The problem was designed to exploit (then) new technologies like social media and crowdsourcing to scale up the recruitment and organization of human work. Panel B illustrates the winning team's strategy for incentivizing team-member recruitment inspired by Ref.~\cite{dodds2003experimental}. Discovering a balloon earned a \$2000 share of the total prize sum, while recruiting someone earned half of that person's award~\cite{tang2011reflecting}.}
\label{fig:red_balloon}
\end{SCfigure*}

\subsection{Crowdsourcing}

Another data source (in addition to digital traces) that widespread online services made possible was \textit{crowdsourcing}---the mobilization and coordination of people through digital platforms. Crowdsourcing has roots in optimization by competition, like Axelrod's computer tournaments (to find the optimal strategy for the iterated Prisoner's Dilemma~\cite{axelrod1984}) and later data mining and machine learning contests. 

A notable type of crowdsourcing challenge involved planning and implementing social mobilization. The most prominent example in this category was DARPA's Red Balloon Challenge, in which teams attempted to locate ten balloons across the continental US. The winning team used a hierarchical system to distribute the award between people actually reporting the locations of the balloons and those recruiting the reporters~\cite{tang2011reflecting,pickard2011time} (see Fig.~\ref{fig:red_balloon}).

An even more influential form of crowdsourcing-based science was the use of Amazon Mechanical Turk (MTurk)~\cite{turk1,turk2}. Designed to recruit and pay workers for small computer-reportable tasks, originally motivated to build training data for image classification, it found use in the social and behavioral sciences as a platform for surveys and experiments. Even though MTurk studies have always been criticized~\cite{mturk_sux}---mostly for uncontrolled biases stemming from the workforce's anonymity---they became very popular, with over a thousand papers per year by the mid-2010s~\cite{turk2}.

\subsection{Science of prediction}

The era of big data brought the insight that AI could predict (at least some kinds of) human behavior extremely well, better even than the theories of domain experts. Even if much could be said about the role of prediction in the scientific endeavor~\cite{toulmin}, many saw this as a call to reform the scientific method. This sentiment was accentuated by the replication crisis in the behavioral sciences~\cite{replication_crisis}, so by the mid-2010s, change was very much in the air. The more extreme voices foresaw an ``end of theory''~\cite{end_of_theory}, where computer predictions would replace theory building (in all senses of the word). More moderate methodologists, however, took the rising AI prowess as an impetus to reform the methodology of the social and behavioral sciences, with computational social science leading the way~\cite{grimmer_roberts_stewart,hofman2017prediction}.

Fast forward to the present and this debate is still reverberating~\cite{hofman2021integrating,hofman2017prediction}, and, yes, there has been a revolution---especially in econometrics, but also in pockets of public health, political science, psychology, and sociology---but by the widespread adoption of \textit{causal inference} methods~\cite{hernan2020causal}, rather than validating theories by their ability to predict. While this development surely has raised the awareness and quality of the scientific explanations coming out of these fields, some of the problems pointed out a decade earlier~\cite{ward2010perils} remain, but have fallen out of focus.

Finally, note that prioritizing prediction over understanding (or theorizing) is not limited to applications of ML. For example, the big-data era also saw a shift towards prediction as a criterion for success in discrete choice modeling (in fields such as econometrics and marketing)~\cite{train2009discrete,rossi2005bayesian}.

\section{The age of generative AI}

When the release of ChatGPT in November 2022 brought an AI chatbot capable of assisting people in their work and everyday lives~\cite{turing_test,biever}, much of the futurism of the preceding age of big data gave way to something more familiar. A chatbot itself was nothing new: the first successful one was Eliza from 1966, and two years later \textsc{Parry} was claimed to have passed the Turing test~\cite{chatbots,weizenbaum1976computer}. However, ChatGPT not only sounded like a human; it also said useful things in a wide variety of situations. It thus made the AI's capabilities accessible, and ostensibly less like interacting with a black box of superhuman predictive powers~\cite{science_in_action} (although it was). This opened an entire orchard of low-hanging fruit as social and behavioral scientists could simply replace humans with chatbots in their favorite experiments (or surveys, or crowdsourcing tasks), and triggered a frenzy that, as of now, shows no sign of dissipating. Furthermore, tasks such as text analysis and image classification became much more accessible. It is not the intention of this paper to review the literature, but we will sketch some trends in this still-young era.

\subsection{Technological developments}

From 2017 onward, AI research and industry witnessed rapid growth, driven predominantly by transformer-based models and generative AI. The Generative Pre-trained Transformer (GPT) series of models showed substantial improvements in natural language understanding and generation~\cite{radford2018gpt}. The Bidirectional Encoder Representations from Transformers (BERT) introduced contextual embeddings, thereby greatly advancing NLP tasks such as question answering and sentiment analysis~\cite{devlin2018bert}. These models are the basis for the LLMs that power the text generation in today's AI agents.

Another milestone was Generative Adversarial Networks (GANs), which demonstrated hitherto unseen levels of realism in image synthesis, style transfer, and data augmentation~\cite{karras2019gan}. Similar success was achieved by so-called diffusion models~\cite{ho2020denoising}, which have proven more stable and are more commonly used today for text-to-X (images, video clips, sound, etc.). At the time of writing, LLMs and diffusion models are combined in commercial applications to make the AI multimodal.

Generative AI holds the distinction of being the first type of successful ML platform to violate the ``garbage in, garbage out'' dictum since they are capable of generating non-factual output even when trained on validated data. This is because their memories are reconstructed on the fly from distributed bits of information rather than retrieved from a fixed storage~\cite{aaai2025presidentialpanel}. Indeed, the earliest LLMs made an impact with their ability to generate coherent but wholly imaginary stories~\cite{radford2019language}. An increased factuality has later been imposed on them by fine-tuning on curated training data~\cite{devlin2018bert,align}.

Market analysts currently believe that generative AI is becoming profitable~\cite{genai_potential}. On the other hand, ever since the launch of ChatGPT, some commentators have seen omens of an approaching plateau in the AI's capabilities, despite the enormous investments~\cite{aaai2025presidentialpanel,shumailov2024ai}. Several commentators also raise concerns about their resource demands, unfair use of training data, deprivation of self-expression, lack of accountability, etc.

\begin{SCfigure*}[0.5]
\centering\includegraphics[width=1.4\linewidth]{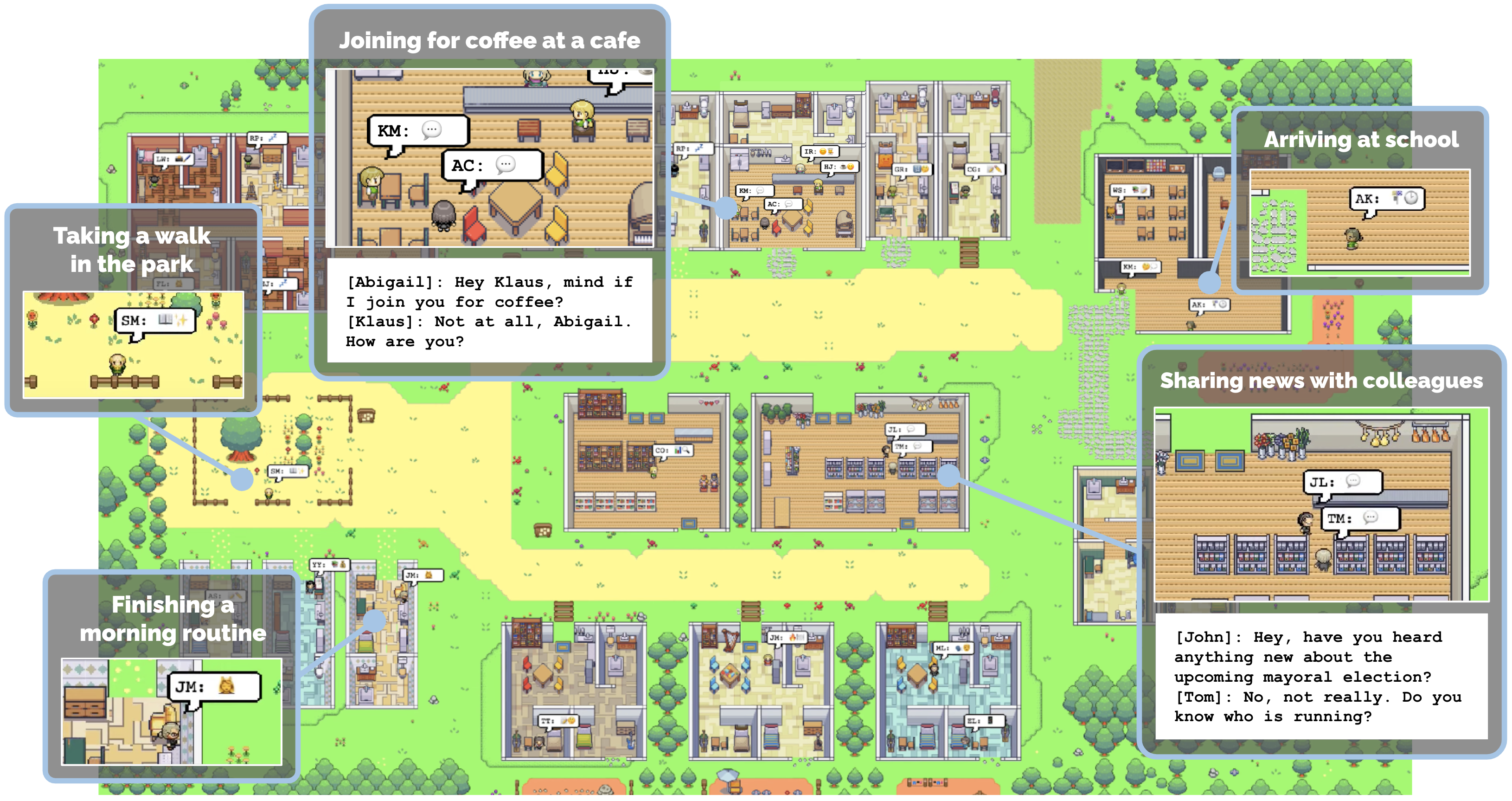}
\caption{\textbf{An illustration of the LLM-chatbot-driven simulations of Ref.~\cite{bernstein}} where 25 agents simulated the inhabitants of a village. Each agent was seeded with a description of their background and their relationships with other agents. The authors observed several aspects of emergent sociality, such as the coordinated organization of a Valentine's Day party. (Reprinted with permission from the authors.)}
\label{fig:generative_agents}
\end{SCfigure*}

\subsection{Active topics}

We are approaching the present and thus the end of the scope of this paper. For the sake of the reader, we will sketch the situation as of writing.

\paragraph{Replacing study objects with chatbots}
The first, and to some extent the least creative, type of sociobehavioral study employing AI chatbots is simply to have them replace humans in experiments or surveys~\cite{horton2023}. These types of studies primarily have a value for documenting the capabilities of AI at the time of the study~\cite{llm_personality,robustness_audit}. They could also serve as a basis for future benchmarking or potentially replace pilot studies in experimental designs~\cite{evans,bail,ziems2023can}. The face value of these studies may seem minor. After all, the actual study objects are not human. But then again, we are in the chatbot era precisely because AI is good at emulating humans. Standard models of humans, like the all-knowing, perpetually optimizing Homo economicus, seem like an easy match, at least in terms of realism.

In an early paper of this kind, Park \textit{et al.}\ demonstrated that communicating LLM agents could organize an event in collaboration and show other signs of emergent social behavior~\cite{bernstein} (see Fig.~\ref{fig:generative_agents}). A persistent merit of this paper is that it popularized the methodology of multi-agent roleplay in a confined (``sandboxed'') environment. Other papers replicated well-established human behavioral studies with AI: for example, Refs.~\cite{mei2024turing,Fontana_Pierri_Aiello_2025} studied a number of game-theoretic experiments with LLMs to conclude that machines are slightly more cooperative than humans. Other studies, on the other hand, found them less cooperative~\cite{jin}. This reflects the general sentiment at the time of writing: AI can emulate humans well enough to replace them in some intellectual activities, but is easily distinguishable from humans in certain tasks~\cite{tsvetkova2024new,align,align_again}.

\paragraph{Studying human-AI interactions}
If we are to believe the AI industry, we don't have any alternative but a future with a much larger presence of AI. Understanding the emergent social consequences of this development is an important research direction~\cite{tsvetkova_human_machine_networks,PEDRESCHI2025104244,lai_diversity,schroeder2025malicious,burton2024how} and motivates several studies involving AI agents.

A common theme in this discussion has been the exploration of AI's ability to influence people's opinions~\cite{engineering_prosociality}. For example, Costello, Pennycook, and Rand~\cite{costello2024durably} demonstrate that dialogues with AI can cause people to relinquish their belief in conspiracy theories for a considerable period. In other studies, Hackenburg \textit{et al.}~\cite{hackenburg2024evaluating,Hackenburg2025} and Lin \textit{et al.}~\cite{Linetal2025} found that dialogues with LLMs can effectively persuade people on political issues. With a careful study design, one could imagine studying AI-rich futures in the wild, but so far we are only aware of ethically compromised attempts to do so~\cite {science2025unethical} (cf.\ Fig.~\ref{fig:ill}D). As a final example, Tessler \textit{et al.}~\cite{tessler2024ai} found that AI could mediate debates on divisive political issues, guiding people toward consensus.

Another line of research on this theme examines how human behavior changes when interacting or collaborating with machines compared with human-only groups. For example, Makovi \textit{et al.}~\cite{makovi2025rewards} found that human-machine populations need both peer rewards and peer punishments to overcome human anti-machine bias in economic games, whereas K\"obis \textit{et al.}~\cite{kobis2025delegation} investigated dishonesty as a side-effect of delegating tasks to AI.

Finally, several studies have addressed the impact of AI on human creativity and productivity in intellectual teamwork. Some studies with positive conclusions, such as Noy and Zhang~\cite{noy2023experimental} and Brynjolfsson \textit{et al.}~\cite{brynjolfsson2025generative}, found workers to be more content and productive when performing writing tasks with AI assistance than without it. Ref.~\cite{dellacqua2024navigating} came to similar conclusions for AI-supported management consultancy. Conclusions from experiments with AI and creativity, on the other hand, seem task-dependent~\cite{lee2024,meincke2025chatgpt}.

\paragraph{Improving analysis tools}
Foundation models like LLMs are capable of many forms of text and image analyses; maybe not as reliably or with the same interpretability as specialized NLP or image analysis tools, but very accessibly. For example, Bermejo \textit{et al.}~\cite{bermejo2024llms} note that for some classification tasks, LLMs can be more accurate than human annotators. Krugmann and Hartmann~\cite{krugmann2024sentiment} are equally optimistic about LLMs' capabilities for sentiment analysis~\cite{gautam2025survey}, while Refs.~\cite{tornberg} and \cite{eddie_annotation} present the opportunities and risks of using LLMs to analyze text. Furthermore, Karjus~\cite{karjus} lists topic classification, historical event cause analysis, social network inference, and missing-data augmentation as suitable tasks for AI in the social sciences.

\paragraph{Replacing scientists with AI}
Replacing people with chatbots is an idea that has not stopped with experiments and media analysis---several studies have sketched how chatbots and AI agents can replace scientists either for some specific tasks~\cite{wuttke2024,focus_group,engzell_wilmers_2026,nguyen-trung2025} or entirely~\cite{closed_loop,sakana_ai,boiko2023autonomous,wang_disco,gottweis2026}. Among the partial tasks where AI could potentially replace social and behavioral scientists are conversational interviews~\cite{wuttke2024} and moderation in focus groups~\cite{focus_group}. The papers proposing AI scientists covering the entire scientific process are usually focused on natural science or engineering, but are also sufficiently general to seem relevant to the social and behavioral sciences. Some of these papers are more programmatic in nature~\cite{closed_loop}, whereas others have implemented pipelines and proven that automated paper production can pass the scrutiny of human peer review~\cite{sakana_ai,boiko2023autonomous}.

\subsection{The missing topic: Using AI to understand humans}

Despite the surge in research on AI agents and a flood of research-free perspective and opinion papers pointing out the possibilities~\cite{christakis_et_al,evans,bail}, studies focusing on humans \textit{per se} are bewilderingly rare. The majority of AI studies published today can, at best, be seen as pilot studies (or ``robustness audits''~\cite{robustness_audit}) for future investigations into actual social behavior. Even agentic-AI papers whose stated motivation is to understand humans---like Argyle \textit{et al.}~\cite{argyle}, who use experiments with AI agents to validate theories about political persuasions---make the relative performance of various AI setups a take-home message. This is especially surprising given that interpreting human behavior, emotions, etc., to provide machines with actionable input is a mature topic in HCI~\cite{pantic2006human}. Perhaps this fruit is just not as low-hanging as the other ones listed above and is simply biding its time to be picked.

\subsection{Drawbacks}

GenAI-based approaches come with some drawbacks that previous methods did not have. For example, both the rapid development and the inherent stochasticity make LLM studies difficult to replicate. This situation is exacerbated by the lack of a unified methodology~
\cite{vaugrante2023}. How much their abilities to replicate human behavior can transfer from one situation to another is uncertain~\cite{robustness_audit}. Moreover, the guardrails and fine-tuning put in place to prevent LLMs from acting offensively or leaking private information can stop working when the models are customized~\cite{even_when}. Finally, generative AI is now in the process of overwriting the human textual corpora that once served as its training data. This risks deteriorating both future AI~\cite{shumailov2024ai} and human knowledge~\cite{llm_personality}.
 
\begin{figure*}
\centering\includegraphics[width=\linewidth]{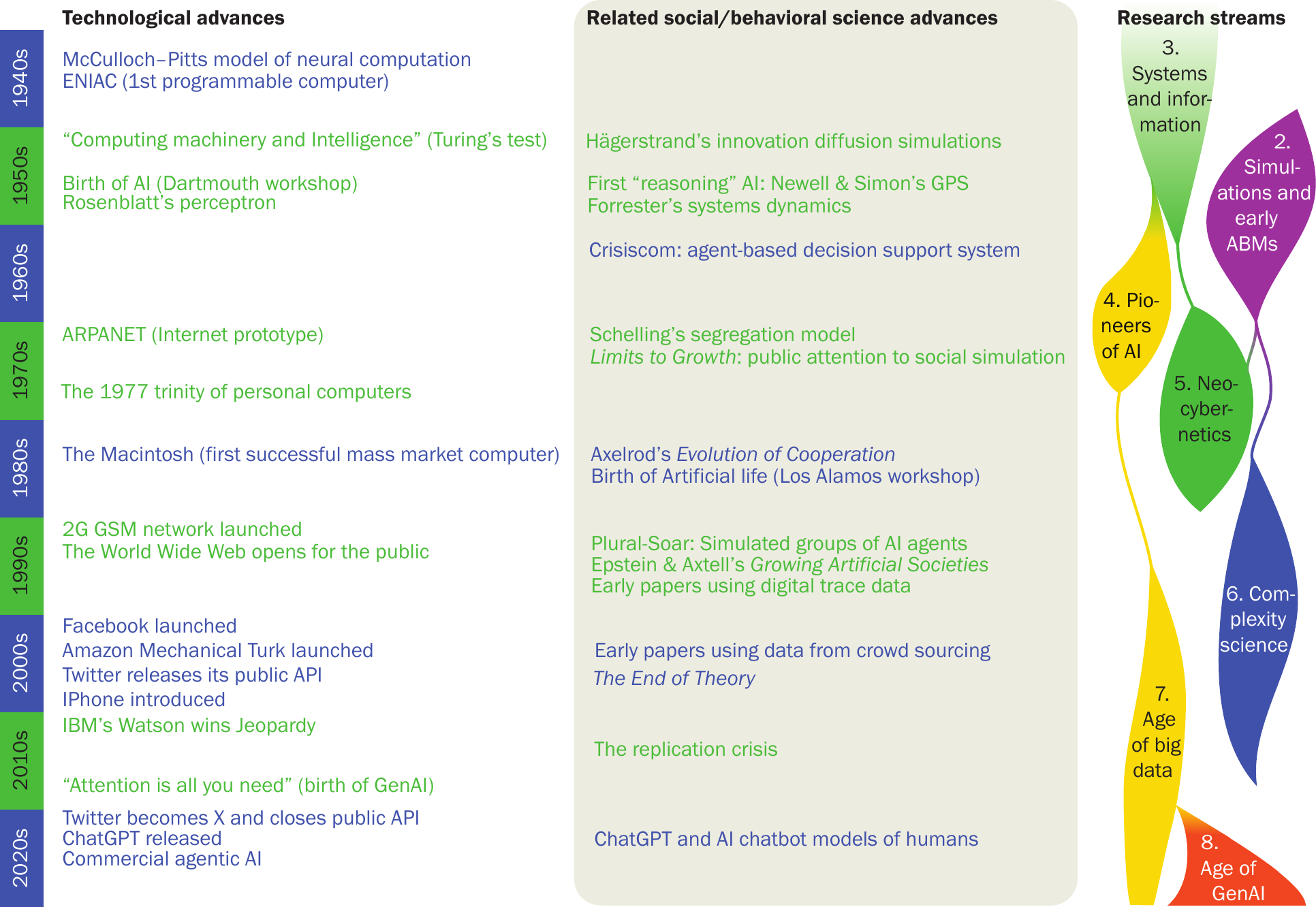}
\caption{\textbf{A timeline of some key moments covered in this paper.} Note that the influence goes in both directions---agent-based modeling of collective intelligence has informed commercial agentic AI, and generative AI has influenced new experiments with chatbots representing humans. The items are all accurate in their order and decade, but within the decade, readability may push items away from their true vertical position. The research streams to the right should be taken with more than a grain of salt. The blobs follow the section structure of the paper and roughly indicate the time periods of the most creative developments. None of these streams has completely died out, and there are practitioners among them who will testify that they are alive and well. Lines between them show important channels of information flow, but there are many missing links and external influences (complexity science was much inspired by chaos theory dynamics at the outset, etc.).}
\label{fig:timeline}
\end{figure*}

\section{Conclusions}

The 75 years leading up to today's agentic AI (Fig.~\ref{fig:timeline}) tell a story of humans and machines being deeply entwined. Maybe it all goes back to us trying to understand ourselves, for what one cannot create, one does not understand (to paraphrase Richard Feynman). From the very first social simulation studies, via the early AI's objective of making machines that learn like children~\cite{weizenbaum1976computer,nilsson}, to today's LLM-based behavioral experiments, the criterion for success has been that machine behavior is indistinguishable from that of humans.

Maybe it is the deeply human fascination with replicas of ourselves and our environment~\cite{simulacra,gelernter} that makes us readily accept simulation as a scientific explanation. Indeed, the methodology of simulation seems to have been in place already before scientists had access to the hardware: recall the 1949 analog simulator of Keynesian economics, the \textsc{Moniac}~\cite{simon1969}, or Helen Abbey's 1951 stochastic simulations by hand of the Reed-Frost model of epidemics~\cite{abbey1952}. Indeed, the highest standard of verification has always been the ability to recreate and forecast what we study~\cite{hofman2017prediction,hofman2021integrating}. This, on the other hand, has not sufficed for computational approaches to become mainstream, like the statistics revolution in the behavioral sciences~\cite{the_cult}. It is tempting to speculate that this is because simulations attempt to explain larger chunks of reality than the typical few causal relations representing the discovery of a sociology or psychology article, or the stylized mathematical models of an economics theory.

Until chess programs reached superhuman levels in the 1990s~\cite{nilsson}, the question of whether machine intelligence should be as good as possible or as human-like as possible made little sense. However, now branches defined by these two goals pull the AI-augmented social and behavioral sciences in different directions. The as-good-as-possible mindset was behind the big data and data-science boom of the 2010s. With new data sets of unprecedented size collected from the digital breadcrumbs we leave behind (thus realizing an old cybernetic dream---cf.\ Fig.~\ref{fig:beer}), ML was able to predict human behavior in ways unattainable to social and behavioral science. Turning this situation into knowledge creation became the \textit{raison d'\^etre} of the new field of computational social science, and spurred an epistemic discourse around the role of prediction in science that is still unfolding~\cite{hofman2017prediction,hofman2021integrating,watts2017solution}. Lately, however, obtaining digital trace data for academic research has become increasingly restricted, ushering in the changes described below.

With the advent of ChatGPT in late 2022 and other easily accessible LLM-based chatbots, the winds changed. The predominant mindset became to harness AI's ability to mimic humans---i.e., to strive for AI to be as aligned with human behavior as possible~\cite{centaur,align,hagendorff2023human}. Naively, this does not seem to require a substantial rethinking of the traditional scientific process. One can simply replace humans with chatbots and proceed with mainstream methods. The catch, of course, is that even though one can align AI and human behavior in some aspects, it is hard to estimate how much this alignment generalizes~\cite{align,align_again,robustness_audit}. Moreover, when it comes to indirect, iterative tasks such as reinforcement learning~\cite{near_optimal} or planning~\cite{cant_plan}, LLMs seem to be less successful in replicating humans. So, in conclusion, generative AI provides a standardized black-box model of humans~\cite{science_in_action} that may be practical and convenient, but it involves many epistemic subtleties~\cite{messeri2024artificial}.

The current fervor of GenAI-fueled social science may subside, but its impact will last. Recent technological progress provides more than just models of humans. We (as in humans and AI combined) now have new ways of processing, summarizing, and structuring data, as well as new methods for generating, systematizing, and disseminating knowledge. Together with the new methods and epistemic perspectives of the big-data era, we now have an abundance of opportunities to understand the social world and human behavior. This will be ever more important since the very same technologies are changing our society and behavior~\cite{postman}, and this change also contributes to our scientific introspection: ``Every new technology allows us to learn something new about ourselves''~\cite{earthquakes}. Social science has always been special in that it is related to what it studies---a ``subject-subject relation''~\cite{giddens}---only now the instruments of inquiry are also part of the equation.

\subsection*{Acknowledgements}
PH was supported by grant no.\ 376435 from the Research Council of Finland and JSPS KAKENHI Grant Number JP 25K01452. MT was supported by ERC grant No.\ 101170272 HUMANET.

\bibliographystyle{abbrv}
\bibliography{refs}

@techreport{manyika2011big,
 author = {Manyika, James and Chui, Michael and Brown, Brad and Bughin, Jacques and Dobbs, Richard and Roxburgh, Charles and Byers, Angela Hung},
 title = {Big Data: The Next Frontier for Innovation, Competition, and Productivity},
 institution = {McKinsey Global Institute},
 year = {2011}
}

@techreport{aaai2025presidentialpanel,
 title = {Report of the {AAAI} 2025 Presidential Panel on the Future of {AI} Research},
 author = {Francesca Rossi and Christian Bessiere and Joydeep Biswas and Rodney Brooks and Vincent Conitzer and Thomas G. Dietterich and Virginia Dignum and Oren Etzioni and Kenneth D. Forbus and Eugene Freuder and Yolanda Gil and Holger Hoos and Eric Horvitz and Subbarao Kambhampati and Henry Kautz and Jihie Kim and Hiroaki Kitano and Alan Mackworth and Karen Myers and Luc De Raedt and Stuart Russell and Bart Selman and Peter Stone and Millind Tambe and Michael Wooldridge and Aditya Akella and Yoshua Bengio and Abeba Birhane and Bill Dally and Fei Fang and Jonathan Gratch and Norm Jouppi and John E. Laird and Amy Luers and Peter Norvig and Besmira Nushi and Balaraman Ravindran and Yoav Shoham and Carles Sierra and Pradeep Varakantham},
 institution = {Association for the Advancement of Artificial Intelligence},
 year = 2025,
 month = mar,
 type = {Technical Report},
 url = {https://www.aaai.org/wp-content/uploads/2025/03/AAAI-2025-PresPanel-Report-Digital-3.7.25.pdf}
}

@book{merleau_ponty,
 author = {Merleau-Ponty, Maurice},
 title = {Phenomenology of Perception},
 year = {1978},
 publisher = {Routledge \& Kegan Paul},
 address = {London}
}

@article{Wallach2018,
 author = {Wallach, Hanna},
 title = {Computational social science $\neq$ computer science + social data},
 journal = {Commun. ACM},
 year = {2018},
 volume = {61},
 number = {3},
 pages = {42-44},
}

@book{hayek1969studies,
 author = {Hayek, Friedrich A.},
 title = {Studies in Philosophy, Politics and Economics},
 year = {1969},
 publisher = {Touchstone},
 address = {New York}
}

@article{shumailov2024ai,
 author = {Shumailov, Ilia and Shumaylov, Zakhar and Zhao, Yiren and Papernot, Nicolas and Anderson, Ross and Gal, Yarin},
 title = {{AI} models collapse when trained on recursively generated data},
 journal = {Nature},
 year = 2024,
 volume = 631,
 pages = {755-759},
}

@techreport{genai_potential,
 title = {The Economic Potential of Generative {AI}: The Next Productivity Frontier},
 author = {Michael Chui and Eric Hazan and Roger Roberts and Alex Singla and Kate Smaje and Alex Sukharevsky and Lareina Yee and Rodney Zemmel},
 institution = {McKinsey \& Company},
 year = 2023,
 month = {June},
}

@inproceedings{krizhevsky2012imagenet,
 author = {Krizhevsky, Alex and Sutskever, Ilya and Hinton, Geoffrey E.},
 title = {{ImageNet} classification with deep convolutional neural networks},
 booktitle = {Advances in Neural Information Processing Systems},
 volume = 25,
 pages = {1097-1105},
 year = 2012
}

@article{lecun2015deep,
 author = {Yann LeCun and Yoshua Bengio and Geoffrey Hinton},
 title = {Deep Learning},
 journal = {Nature},
 year = 2015,
 volume = 521,
 pages = {436-444},
}

@article{lstm,
 author = {S Hochreiter and J Schmidhuber},
 title = {Long Short-Term Memory},
 journal = {Neural Comput.},
 volume = 9,
 number = 8,
 pages = {1735-1780},
 year = 1997,
}

@article{hofman2017prediction,
 author = {Jake M. Hofman and Amit Sharma and Duncan J. Watts},
 title = {Prediction and explanation in social systems},
 journal = {Science},
 year = 2017,
 volume = 355,
 number = 6324,
 pages = {486-488},
}

@book{varela1991embodied,
 author = {Francisco J. Varela and Evan Thompson and Eleanor Rosch},
 title = {The Embodied Mind: Cognitive Science and Human Experience},
 year = 1991,
 publisher = {MIT Press},
 address = {Cambridge MA},
}

@article{new_social_science,
author = {Leslie Henrickson and Bill McKelvey},
title = {Foundations of ``new'' social science: Institutional legitimacy from philosophy, complexity science, postmodernism, and agent-based modeling},
journal = {Proc. Natl. Acad. Sci. USA},
volume = 99,
pages = {7288-7295},
year = 2002,
}

@inproceedings{cant_plan,
author = {Kambhampati, Subbarao and Valmeekam, Karthik and Guan, Lin and Verma, Mudit and Stechly, Kaya and Bhambri, Siddhant and Saldyt, Lucas and Murthy, Anil},
title = {{LLMs} can't plan, but can help planning in {LLM}-modulo frameworks},
year = {2024},
publisher = {JMLR.org},
booktitle = {Proceedings of the 41st International Conference on Machine Learning},
pages = 921,
numpages = 13,
location = {Vienna, Austria},
series = {ICML'24}
}

@article{align_again,
 author = {Tao, Yan and Viberg, Olga and Baker, Ryan S and Kizilcec, Ren\'e F},
 title = {Cultural bias and cultural alignment of large language models},
 journal = {PNAS Nexus},
 volume = 3,
 number = 9,
 pages = {pgae346},
 year = 2024,
}

@inproceedings{align,
author = {Aher, Gati and Arriaga, Rosa I. and Kalai, Adam Tauman},
title = {Using large language models to simulate multiple humans and replicate human subject studies},
year = 2023,
publisher = {JMLR.org},
booktitle = {Proceedings of the 40th International Conference on Machine Learning},
pages = 17,
numpages = 35,
location = {Honolulu, Hawaii, USA},
series = {ICML'23}
}

@unpublished{near_optimal,
 author = {Hao Li and Gengrui Zhang and Petter Holme and Shuyue Hu and Zhen Wang},
 title = {Large Language Models are Near-Optimal Decision-Makers with a Non-Human Learning Behavior},
 note = {Preprint arXiv:2506.16163},
 year = {2025}
}

@unpublished{even_when,
 author = {Xiangyu Qi and Yi Zeng and Tinghao Xie and Pin-Yu Chen and Ruoxi Jia and Prateek Mittal and Peter Henderson},
 title = {Fine-tuning Aligned Language Models Compromises Safety, Even When Users Do Not Intend To!},
 note = {Preprint arXiv:2310.03693},
 year = 2023
}

@unpublished{engzell_wilmers_2026,
title={The paper factory},
note={preprint socarxiv/24xfq\_v2},
author={Engzell, Per and Wilmers, Nathan},
year=2026, 
}

@unpublished{robustness_audit,
 author = {Jinyi Ye and Lei Cao and Ding Chen and Emilio Ferrara},
 title = {Stop Drawing Scientific Claims from {LLM} Social Simulations Without Robustness Audits},
 note = {Preprint arXiv:2605.18890},
 year = 2026
}

@inproceedings{simonyan2014very,
  author    = {Simonyan, Karen and Zisserman, Andrew},
  title     = {Very deep convolutional networks for large-scale image recognition},
  booktitle = {Proceedings of the 3rd International Conference on Learning Representations},
  year      = {2015}
}

@article{hendershott,
author = {HENDERSHOTT, TERRENCE and JONES, CHARLES M. and MENKVELD, ALBERT J.},
title = {Does Algorithmic Trading Improve Liquidity?},
journal = {J. Finance},
volume = 66,
number = 1,
pages = {1-33},
year = 2011
}

@inproceedings{sushil_tweet,
author = {Chu, Zi and Gianvecchio, Steven and Wang, Haining and Jajodia, Sushil},
title = {Who is tweeting on {T}witter: human, bot, or cyborg?},
year = 2010,
booktitle = {Proceedings of the 26th Annual Computer Security Applications Conference},
pages = {21–30},
}

@inproceedings{kwak2010twitter,
  title={What is {T}witter, a social network or a news media?},
  author={Kwak, Haewoon and Lee, Changhyun and Park, Hosung and Moon, Sue},
  booktitle={Proceedings of the 19th international conference on World wide web},
  pages={591-600},
  year=2010
}

@unpublished{eddie_annotation,
 author = {Joachim Baumann and Paul R\"ottger and Aleksandra Urman and Albert Wendsj\"o and Flor Miriam Plaza-del-Arco and Johannes B. Gruber and Dirk Hovy},
 title = {Large Language Model Hacking: Quantifying the Hidden Risks of Using {LLMs} for Text Annotation},
 note = {Preprint arXiv:2509.08825},
 year = 2025
}

@inproceedings{szegedy2015going,
 author = {Szegedy, Christian and Liu, Wei and Jia, Yangqing and Sermanet, Pierre and Reed, Scott and Anguelov, Dragomir and Erhan, Dumitru and Vanhoucke, Vincent and Rabinovich, Andrew},
 title = {Going deeper with convolutions},
 booktitle = {Proceedings of the IEEE Conference on Computer Vision and Pattern Recognition},
 pages = {1-9},
 year = {2015}
}

@inproceedings{he2016deep,
 author = {He, Kaiming and Zhang, Xiangyu and Ren, Shaoqing and Sun, Jian},
 title = {Deep residual learning for image recognition},
 booktitle = {Proceedings of the IEEE Conference on Computer Vision and Pattern Recognition},
 pages = {770-778},
 year = {2016}
}

@inproceedings{graves2013speech,
 author = {Graves, Alex and Mohamed, Abdel-Rahman and Hinton, Geoffrey},
 title = {Speech recognition with deep recurrent neural networks},
 booktitle = {IEEE International Conference on Acoustics, Speech and Signal Processing},
 pages = {6645-6649},
 year = {2013}
}

@article{krugmann2024sentiment,
 author = {Krugmann, Jan Ole and Hartmann, Jochen},
 title = {Sentiment Analysis in the Age of Generative {AI}},
 journal = {Cust. Needs Solut.},
 year = 2024,
 volume = 11,
 pages = 3,
}

@article{boiko2023autonomous,
 author = {Boiko, Daniil A. and MacKnight, Robert and Kline, Ben and Gomes, Gabe},
 title = {Autonomous chemical research with large language models},
 journal = {Nature},
 year = 2023,
 volume = 624,
 pages = {570-578},
}

@inproceedings{mikolov2013efficient,
  author    = {Mikolov, Tomas and Chen, Kai and Corrado, Greg and Dean, Jeffrey},
  title     = {Efficient Estimation of Word Representations in Vector Space},
  booktitle = {Proceedings of the 1st International Conference on Learning Representations},
  year      = {2013}
}

@unpublished{tornberg,
 author = {Petter T\"ornberg},
 title = {How to use {LLMs} for Text Analysis},
 note = {Preprint arXiv:2307.13106},
 year = {2023}
}

@inproceedings{attention,
 author = {Vaswani, Ashish and Shazeer, Noam and Parmar, Niki and Uszkoreit, Jakob and Jones, Llion and Gomez, Aidan N. and Kaiser, Lukasz and Polosukhin, Illia},
 title = {Attention Is All You Need},
 booktitle = {Advances in Neural Information Processing Systems},
 volume = 30,
 pages = {5998-6008},
 year = {2017}
}

@incollection{diplo,
 author = {P Holme and G Ghoshal},
 title = {The diplomat's dilemma: Maximal power for minimal effort in social networks},
 booktitle = {Adaptive Networks: Theory, Models and Applications},
 publisher = {Springer},
 address = {Berlin},
 year = 2009,
 editor = {T Gross and H Sayama},
pages = {269-288}
}

@incollection{coming_evolution,
 author = {J. Doyne Farmer and Alletta d'A. Belin},
 title = "Artificial Life: The Coming Evolution",
 booktitle = "Artificial Life II",
 publisher = {Addison-Wesley},
 address = {Redwood City CA},
pages = {815-840},
 year = {1992},
 editor = {Christopher G. Langton and C Taylor and J. Doyne Farmer and Steen Rasmussen}
}

@article{watts2017solution,
 author = {Duncan J. Watts},
 title = {Should social science be more solution-oriented?},
 journal = {Nat. Hum. Behav.},
 year = 2017,
 volume = 1,
 number = 1,
 pages = 15,
}

@article{minsky1982why,
 author = {Marvin Minsky},
 title = {Why people think computers can't},
 journal = {AI Mag.},
 year = 1982,
 volume = 3,
 number = 4,
 pages = {3-15},
}

@article{mcluhan,
 title = {The reversal of the overheated image},
 author = {M McLuhan},
 journal = {Playboy},
volume = 15,
 pages = {131-134},
 year = {1968},
}

@article{nonaka2006organizational,
 title = {Organizational Knowledge Creation Theory: Evolutionary Paths and Future Advances},
 author = {Nonaka, Ikujiro and von Krogh, Georg and Voelpel, Sven},
 journal = {Organ. Stud.},
 volume = {27},
 number = {8},
 pages = {1179-1208},
 year = {2006},
}

@ARTICLE {durkin,
author = {J. Durkin},
journal = {IEEE Intell. Syst.},
title = {Expert Systems: A View of the Field},
year = 1996,
volume = 11,
pages = {56-63}
}

@incollection{colby1963,
 author = {K M Colby},
 title = "Computer Simulation of a Neurotic Process",
 booktitle = "Computer Simulation of Personality",
 publisher = {John Wiley \& Sons},
 address = {New York},
 year = 1963,
 editor = {Tompkins, S S and Messick, S},
}

@incollection{feigen,
 author = {E Feigenbaum},
 title = {Computer simulation of human behavior},
 booktitle = {Midwest Human Factors Symposium},
 year = {1963},
 publisher = {AC Spark Plug},
 address = {Chicago},
 pages = {1-8}
}

@incollection{zelditch_evan,
 author = {Morris Zelditch and William M. Evan},
 title = {Simulated bureaucracies: A methodological analysis},
 booktitle = {Simulation in Social Science: Readings},
 editor = {H Guetzkow},
 year = {1962},
 publisher = {Prentice-Hall},
 address = {Englewood Cliffs NJ},
 pages = {48-60}
}

@article{nowak_may,
 title = "Evolutionary games and spatial chaos",
 author = "M A Nowak and R M May",
 year = 1982,
journal = "Nature",
volume = 359,
pages = {826–829},
}

@article{machine_behavior,
 author = {Iyad Rahwan and Manuel Cebrian and Nick Obradovich and Josh Bongard and Jean-François Bonnefon and Cynthia Breazeal and Jacob W. Crandall and Nicholas A. Christakis and Iain D. Couzin and Matthew O. Jackson and Nicholas R. Jennings and Ece Kamar and Isabel M. Kloumann and Hugo Larochelle and David Lazer and Richard McElreath and Alan Mislove and David C. Parkes and Alex Sandy Pentland and Margaret E. Roberts and Azim Shariff and Joshua B. Tenenbaum and Michael Wellman},
 title = {Machine behaviour},
 year = {2019},
 journal = {Nature},
 volume = 568,
 pages = {477–486}
}

@article{wang_disco,
 author = {Hanchen Wang and Tianfan Fu and Yuanqi Du and Wenhao Gao and Kexin Huang and Ziming Liu and Payal Chandak and Shengchao Liu and Peter Van Katwyk and Andreea Deac and Anima Anandkumar and Karianne Bergen and Carla P. Gomes and Shirley Ho and Pushmeet Kohli and Joan Lasenby and Jure Leskovec and Tie-Yan Liu and Arjun Manrai and Debora Marks and Bharath Ramsundar and Le Song and Jimeng Sun and Jian Tang and Petar Veličković and Max Welling and Linfeng Zhang and Connor W. Coley and Yoshua Bengio and Marinka Zitnik},
 title = {Scientific discovery in the age of artificial intelligence},
 journal = {Nature},
 volume = 620,
 pages = {47–60},
 year = 2023}

@book{rashevsky,
 title = "Mathematical Theory of Human Relations",
 author = "Rashevsky, N",
 year = 1947,
publisher = {Principia Press},
address = {Bloomington IN}
}

@book{postman,
 title = "Technopoly",
 author = "Postman, N",
 year = 1993,
publisher = {Alfred A. Knopf},
address = {New York}
}

@article{nk_model,
title = {Towards a general theory of adaptive walks on rugged landscapes},
journal = {J. Theor. Biol.},
volume = {128},
number = {1},
pages = {11-45},
year = {1987},
author = {Stuart Kauffman and Simon Levin},
}

@article{FRENKEN2000257,
title = {A complexity approach to innovation networks. The case of the aircraft industry (1909–1997)},
journal = {Res. Policy},
volume = {29},
number = {2},
pages = {257-272},
year = {2000},
author = {Koen Frenken},
}

@article{levinthal,
author = {Levinthal, Daniel A.},
title = {Adaptation on Rugged Landscapes},
journal = {Manag. Sci.},
volume = {43},
number = {7},
pages = {934-950},
year = {1997},
}

@book{gelernter,
 title = {Mirror Worlds},
 author = {Gelernter, David Hillel},
 year = 1991,
 publisher = {Oxford University Press},
 address = {New York},
}

@book{bara,
 title = {Network Science},
 author = {A-L Barab\'asi},
 year = 2016,
 publisher = {Cambridge University Press},
 address = {Cambridge},
}

@book{wasserman1994social,
 title = {Social Network Analysis: Methods and Applications},
 author = {Wasserman, Stanley and Faust, Katherine},
 year = 1994,
 publisher = {Cambridge University Press},
 address = {Cambridge},
}

@article{arthur1989competing,
 author = {W. Brian Arthur},
 title = {Competing Technologies, Increasing Returns, and Lock-In by Historical Events},
 journal = {Econ. J.},
 year = {1989},
 volume = {99},
 number = {394},
 pages = {116-131},
}

@article{jennings_moreno,
 title = "Statistics of social configurations",
 author = "J L Moreno and H H Jennings",
 year = 1938,
journal = "Sociometry",
volume = 1,
pages = {342-374},
}

@techreport{radford2019language,
 author = {Radford, Alec and Wu, Jeffrey and Child, Rewon and Luan, David and Amodei, Dario and Sutskever, Ilya},
 title = {Language Models are Unsupervised Multitask Learners},
 institution = {OpenAI},
 year = {2019},
 note = {OpenAI Technical Report},
 url = {https://cdn.openai.com/better-language-models/language_models_are_unsupervised_multitask_learners.pdf}
}

@inproceedings{sun2017revisiting,
 title = {Revisiting unreasonable effectiveness of data in deep learning era},
 author = {Sun, Chen and Shrivastava, Abhinav and Singh, Saurabh and Gupta, Abhinav},
 booktitle = {Proceedings of the IEEE International Conference on Computer Vision},
 pages = {843-852},
 year = {2017}
}

@article{halevy2009unreasonable,
 title = {The unreasonable effectiveness of data},
 author = {Halevy, Alon and Norvig, Peter and Pereira, Fernando},
 journal = {IEEE Intell. Syst.},
 volume = 24,
 number = 2,
 pages = {8-12},
 year = 2009,
 publisher = {IEEE}
}

@book{the_cult,
 title = {The Cult of Statistical Significance},
 author = {S T Ziliak and D N McCloskey},
 year = 2008,
 publisher = {University of Michigan Press},
address = {Ann Arbor}
}

@article{hinton_deep_learning,
author = {Hinton, Geoffrey E. and Osindero, Simon and Teh, Yee-Whye},
title = {A fast learning algorithm for deep belief nets},
year = {2006},
publisher = {MIT Press},
address = {Cambridge MA},
volume = 18,
number = 7,
journal = {Neural Comput.},
pages = {1527–1554},
}

@article{rosenblatt_perceptron,
 title = "The perceptron: a probabilistic model for information storage and organization in the brain",
 author = "F Rosenblatt",
 year = 1958,
journal = "Psychol. Rev.",
volume = 65,
pages = {386-408},
}

@book{morin,
 title = "Method 1: The Nature of Nature",
 author = "E Morin",
 year = 1992,
 publisher = "Peter Lang",
 address = "New York"
}

@book{heims,
 title = "The Cybernetics Group",
 author = "S J Heims",
 year = 1991,
 publisher = "MIT Press",
 address = "Cambridge MA"
}

@book{mccorduck,
 title = "Machines Who Think",
 author = "Pamela McCorduck",
 year = 1991,
 publisher = "A K Peters",
 address = "Natick MA"
}

@book{hayek,
 title = "The Sensory Order",
 author = "F A Hayek",
 year = 1952,
 publisher = "University of Chicago Press",
 address = "Chicago"
}

@book{macy_conference,
 title = "Cybernetics: {The Macy} Conferences 1946-1953",
 author = "C Pias",
 year = 2016,
 publisher = "University of Chicago Press",
 address = "Chicago"
}

@book{wiener_human_use,
 title = "The Human Use of Human Beings",
 author = "N Wiener",
 year = 1950,
 publisher = "Houghton Mifflin",
 address = "Boston"
}

@book{cybernetic_moment,
 title = "The Cybernetic Moment",
 author = "R R Kline",
 year = 2015,
 publisher = "Johns Hopkins University Press",
 address = "Baltimore"
}

@book{merton,
 title = "Social Theory and Social Structure",
 author = "R K Merton",
 year = 1949,
 publisher = "The Free Press",
 address = "Glencoe IL"
}

@book{toulmin,
 title = "Foresight and Understanding",
 author = "S Toulmin",
 year = 1963,
 publisher = "Harper \& Row",
 address = "New York"
}

@book{parsons,
 title = "The Social System",
 author = "T Parsons",
 year = 1951,
 publisher = "The Free Press",
 address = "Glencoe IL"
}

@article{degrowth,
 author = "Kallis, Giorgos and Kostakis, Vasilis and Lange, Steffen and Muraca, Barbara and Paulson, Susan and Schmelzer MAtthias",
 title = "Research On Degrowth", 
 journal = "Annu. Rev. Environ. Resour.",
 year = 2018,
 volume = 43,
 pages = "291-316",
 }

@article{feldman_ballard, title = {Connectionist models and their properties}, volume = 6, journal = {Cogn. Sci.}, author = {J H Feldman and D H Ballard}, year = 1982, pages = {205-254}}

@book{lyon,
 title = "Surveillance After Snowden",
 author = "D Lyon",
 year = 2015,
 publisher = "Polity Press",
 address = "Malden MA"
}

@book{medina,
 title = "Cybernetic Revolutionaries: Technology and Politics in {Allende's Chile}",
 author = "E Medina",
 year = 2011,
 publisher = "MIT Press",
 address = "Cambridge MA"
}

@book{beer,
 title = "A Platform for Change",
 author = "S Beer",
 year = 1975,
 publisher = "John Wiley \& Sons",
 address = "New York"
}

@book{odum,
 title = "Environment, Power and Society",
 author = "H T Odum",
 year = 1971,
 publisher = "John Wiley \& Sons",
 address = "New York"
}

@book{pickering,
 title = "The Cybernetic Brain",
 author = "A Pickering",
 year = 2011,
 publisher = "The University of Chicago Press",
 address = "Chicago"
}

@book{mayor,
 title = "Gods and Robots: Myths Machines, and Ancient Dreams of Technologys",
 author = "A Mayor",
 year = 2018,
 publisher = "Princeton University Press",
 address = "Princeton NJ"
}

@book{hegemony,
 title = "The Hegemony of Growth",
 author = "M Schmelzer",
 year = 2016,
 publisher = "Cambridge University Press",
 address = "Cambridge"
}

@article{mcculloch_pitts,
 author = {Warren McCulloch and Walter Pitts},
 title = {A logical calculus of ideas immanent in nervous activity},
 journal = {Bull. Math. Biophys.},
 volume = 5,
 pages = {115–133},
 year = 1943
}

@article{turing,
 author = {Turing, A. M.},
 title = "{Computing machinery and intelligence}",
 journal = {Mind},
 volume = 59,
 pages = {433-460},
 year = 1950
}

@book{PfeiferScheier1999,
 author = {Pfeifer, Rolf and Scheier, Christian},
 title = {Understanding Intelligence},
 publisher = {MIT Press},
address = {Cambridge MA},
 year = 1999,
}

@article{gershenson_sayama,
 author = {Gershenson, Carlos and Trianni, Vito and Werfel, Justin and Sayama, Hiroki},
 title = {Self-Organization and Artificial Life},
 journal = {Artif. Life},
 volume = {26},
 number = {3},
 pages = {391-408},
 year = {2020},
}

@unpublished{llm_personality,
 author = {Greg Serapio-Garc{\'\i}a and Mustafa Safdari and Clément Crepy and Luning Sun and Stephen Fitz and Peter Romero and Marwa Abdulhai and Aleksandra Faust and Maja Matari\'c},
 title = {Personality traits in large language models},
 year = 2023,
 note = {preprint arxiv:2307.00184},
}

@unpublished{yakura_language,
 author = {Hiromu Yakura and Ezequiel Lopez-Lopez and Levin Brinkmann and Ignacio Serna and Prateek Gupta and Iyad Rahwan},
 title = {Empirical evidence of {Large Language Model}'s influence on human spoken communication},
 year = 2024,
 note = {preprint arxiv:2409.01754},
}

@unpublished{lai_diversity,
 author = {Shiyang Lai and Yujin Potter and Junsol Kim and Richard Zhuang and Dawn Song and James Evans},
 title = {Evolving {AI} Collectives to Enhance Human Diversity and Enable Self-Regulation},
 year = 2024,
 note = {preprint arxiv:2402.12590},
}

@unpublished{holme_arxiv,
 author = {P Holme},
 title = {What complexity science is, and why},
 year = {2022},
 note = {preprint arxiv:2201.03762},
}

@article{forrester_industrial,
 author = {J W Forrester},
 title = {Industrial dynamics},
 year = {1958},
 journal = {Harv. Bus. Rev.},
 volume = 16,
 pages = {37-66}
}

@article{lazer2009computational,
 title = {Computational social science},
 author = {Lazer, David and Pentland, Alex and Adamic, Lada and Aral, Sinan and Barab\'asi, Albert-László and Brewer, Devon and Christakis, Nicholas and Contractor, Noshir and Fowler, James and Gutmann, Myron and Jebara, Tony and King, Gary and Macy, Michael and Roy, Deb and Van Alstyne, Marshall},
 journal = {Science},
 volume = {323},
 number = {5915},
 pages = {721-723},
 year = {2009},
}

@book{earthquakes,
 author = {S Basar and D Coupland and H U Obrist},
 title = {The Age of Earthquakes: A Guide to the Extreme Present},
 publisher = {Penguin Books},
 year = 2015,
 address = {London},
}

@book{salganik2017bit,
 author = {Salganik, Matthew J.},
 title = {Bit by Bit: Social Research in the Digital Age},
 publisher = {Princeton University Press},
 year = 2017,
 address = {Princeton NJ},
}

@article{rwb,
 author = {Arturo Rosenblueth and Norbert Wiener and Julian Bigelow},
 title = {Behavior, purpose and teleology},
 year = {1943},
 journal = {Philos. Sci.},
 volume = 10,
 pages = {18-24}
}

@article{holme_lilje,
 author = {P Holme and F Liljeros},
 title = {Mechanistic models in computational social science},
 year = 2015,
 journal = {Front. Phys.},
 volume = 3,
 pages = 78
}

@unpublished{closed_loop,
title = {The Future of Fundamental Science Led by Generative Closed-Loop Artificial Intelligence},
author = {Hector Zenil and Jesper Tegn\'{e}r and Felipe S. {Abrah\~{a}o} and Alexander Lavin and Vipin Kumar and Jeremy G. Frey and Adrian Weller and Larisa Soldatova and Alan R. Bundy and Nicholas R. Jennings and Koichi Takahashi and Lawrence Hunter and Saso Dzeroski and Andrew Briggs and Frederick D. Gregory and Carla P. Gomes and Jon Rowe and James Evans and Hiroaki Kitano and Ross King},
year = 2023,
note = {Preprint arXiv:2307.07522}
}

@article{huberman_www,
author = {Bernardo A. Huberman and Peter L. T. Pirolli and James E. Pitkow and Rajan M. Lukose },
title = {Strong Regularities in {World Wide Web} Surfing},
journal = {Science},
volume = 280,
number = 5360,
pages = {95-97},
year = 1998,
}

@book{train2009discrete,
 author = {Train, Kenneth E.},
 title = {Discrete Choice Methods with Simulation},
 edition = {2},
 year = {2009},
 publisher = {Cambridge University Press},
 address = {Cambridge},
}

@book{rossi2005bayesian,
 author = {Rossi, Peter E. and Allenby, Greg M. and McCulloch, Robert},
 title = {Bayesian Statistics and Marketing},
 year = {2005},
 publisher = {John Wiley \& Sons},
 series = {Wiley Series in Probability and Statistics},
 address = {Chichester}
}

@unpublished{turing_test, 
 author		 = "C R Jones and B K Bergen",
 title			 = "Large Language Models Pass the {T}uring Test", 
 year			 = 2025,
 note			 = "Preprint arXiv:2503.23674"
}

@article{biever,
 author = {C Biever},
 title = {{ChatGPT} broke the {Turing} test: the race is on for new ways to assess {AI}},
 year = 2023,
 journal = {Nature},
 volume = 619,
 pages = {686-689}
}

@unpublished{radford2018gpt,
 author = {Radford, Alec and Narasimhan, Karthik and Salimans, Tim and Sutskever, Ilya},
 title = {Improving Language Understanding by Generative Pre-Training},
 year = {2018},
 note = {OpenAI Technical Report}
}

@unpublished{devlin2018bert,
 author = {Devlin, Jacob and Chang, Ming-Wei and Lee, Kenton and Toutanova, Kristina},
 title = {{BERT}: Pre-training of Deep Bidirectional Transformers for Language Understanding},
 note = {Preprint arXiv:1810.04805},
 year = 2018
}

@inproceedings{karras2019gan,
 author = {Karras, Tero and Laine, Samuli and Aila, Timo},
 title = {A Style-Based Generator Architecture for Generative Adversarial Networks},
 booktitle = {Proceedings of the IEEE Conference on Computer Vision and Pattern Recognition},
 pages = {4401-4410},
 year = 2019
}

@article{ho2020denoising,
 author = {Ho, Jonathan and Jain, Ajay and Abbeel, Pieter},
 title = {Denoising Diffusion Probabilistic Models},
 journal = {Adv. Neural Inf. Process.},
 volume = {33},
 pages = {6840-6851},
 year = {2020}
}

@unpublished{sakana_ai,
 author = {Chris Lu and Cong Lu and Robert Tjarko Lange and Jakob Foerster and Jeff Clune and David Ha},
 title = {The {AI} scientist: Towards fully automated open-ended scientific discovery},
 year = {2024},
 note = {Preprint arXiv:2408.06292},
}

@article{jin,
 author = {Han, Jin AND Battu, Balaraju AND Romić, Ivan AND Rahwan, Talal AND Holme, Petter},
 journal = {PLOS ONE},
 title = {Static network structure cannot stabilize cooperation among large language model agents},
 year = {2025},
 volume = {20},
 pages = {0320094},
 number = {5},
}

@article{Fontana_Pierri_Aiello_2025, title = {Nicer than Humans: How Do Large Language Models Behave in the Prisoner’s Dilemma?}, volume = {19}, number = {1}, journal = {Proc. Int. AAAI Conf. Weblogs. Soc. Media.}, author = {Fontana, Nicolò and Pierri, Francesco and Aiello, Luca Maria}, year = {2025}, month = {Jun.}, pages = {522-535} }

@inproceedings{bernstein,
author = {Park, Joon Sung and O'Brien, Joseph and Cai, Carrie Jun and Morris, Meredith Ringel and Liang, Percy and Bernstein, Michael S.},
title = {Generative Agents: Interactive Simulacra of Human Behavior},
year = {2023},
publisher = {Association for Computing Machinery},
address = {New York},
booktitle = {Proceedings of the 36th Annual ACM Symposium on User Interface Software and Technology},
pages = {2},
numpages = 22,
}

@article{fehr2000cooperation,
 author = {Fehr, Ernst and G{\"a}chter, Simon},
 title = {Cooperation and Punishment in Public Goods Experiments},
 journal = {Am. Econ. Rev.},
 year = {2000},
 volume = {90},
 number = {4},
 pages = {980-994},
}

@article{PEDRESCHI2025104244,
title = {Human-{AI} coevolution},
journal = {Artif. Intell.},
volume = {339},
pages = {104244},
year = {2025},
author = {Dino Pedreschi and Luca Pappalardo and Emanuele Ferragina and Ricardo Baeza-Yates and Albert-László Barab\'asi and Frank Dignum and Virginia Dignum and Tina Eliassi-Rad and Fosca Giannotti and János Kert\'esz and Alistair Knott and Yannis Ioannidis and Paul Lukowicz and Andrea Passarella and Alex Sandy Pentland and John Shawe-Taylor and Alessandro Vespignani},
}

@unpublished{bermejo2024llms,
 author = {Vicente J. Bermejo and Andres Gago and Ramiro H. G{\'a}lvez and Nicol{\'a}s Harari},
 title = {{LLMs} Outperform Outsourced Human Coders on Complex Textual Analysis},
 note = {Preprint SSRN/5020034},
 year = {2024},
}

@inproceedings{pantic2006human, 
 title = {Human computing and machine understanding of human behavior: A survey},
 author = {Pantic, Maja and Pentland, Alex and Nijholt, Anton and Huang, Thomas},
 booktitle = {Proceedings of the 8th international conference on Multimodal interfaces},
 pages = {239-248},
 year = {2006}
}

@article{hedstrom_abm,
author = {Peter Hedstr\"om and Gianluca Manzo},
title = {Recent Trends in Agent-based Computational Research: A Brief Introduction},
journal = {Sociol. Methods Res.},
volume = 44,
number = 2,
pages = {179-185},
year = 2015,
}

@article{bail,
author = {Christopher A. Bail},
title = {Can Generative {AI} improve social science?},
journal = {Proc. Natl. Acad. Sci. USA},
volume = {121},
number = {21},
pages = {e2314021121},
year = {2024},
}

@article{christakis_et_al,
author = {Igor Grossmann and Matthew Feinberg and Dawn C. Parker and Nicholas A. Christakis and Philip E. Tetlock and William A. Cunningham },
title = {{AI} and the transformation of social science research},
journal = {Science},
volume = 380,
pages = {1108-1109},
year = 2023,}

@article{evans,
author = {J Sourati and J Evans},
year = 2023,
volume = 7,
title = {Accelerating science with human-aware artificial intelligence},
pages = {1682–1696},
journal = {Nat. Hum. Behav.}}

@article{chatbots,
title = {From {Eliza} to {XiaoIce}: challenges and opportunities
with social chatbots},
author = {Heung-Yeung Shum and Xiao-Dong He and Di Li},
journal = {Front. Inform. Technol. Electron. Eng.},
year = 2018,
volume = 19,
number = 1,
pages = {10-26}}

@book{maynard_smith,
title = {Evolution and the Theory of Games},
author = {J {Maynard Smith}},
year = 1982,
publisher = {Cambridge University Press},
address = {Cambridge}}

@article{wolfers2004prediction,
 title = {Prediction markets},
 author = {Wolfers, Justin and Zitzewitz, Eric},
 journal = {J. Econ. Perspect.},
 volume = {18},
 number = {2},
 pages = {107-126},
 year = {2004},
}

@book{surfing,
title = {Surfing the Edge of Chaos},
author = {Richard Pascale and Mark Milleman and Linda Gioja},
year = 2001,
publisher = {Crown},
address = {New York}}

@book{scott_page,
title = {Complex Adaptive Systems},
author = {S E Page},
year = 2007,
publisher = {Princeton University Press},
address = {Princeton NJ}}

@techreport{abm_economics,
 author = {Axtell, Robert and Farmer, J Doyne},
 title = {Agent-based modeling in economics and finance: Past, present, and future},
 institution = {Oxford Institute for New Economic Thinking},
 number = {2022-10},
 year = 2022
}

@book{abm_business,
title = {Managing Business Complexity},
author = {M J North and C M Macal},
year = 2007,
publisher = {Oxford University Press},
address = {Oxford}}

@book{stephen_levy,
title = {Artificial Life},
author = {S Levy},
year = 1993,
publisher = {Penguin Random House},
address = {New York}}

@book{holland,
title = {Emergence},
author = {J H Holland},
year = 1998,
publisher = {Oxford University Press},
address = {Oxford}}

@book{waldrop,
title = {Complexity},
author = {M M Waldrop},
year = 1992,
publisher = {Simon \& Schuster},
address = {New York}}

@book{bett,
title = {Introduction to Urban Science},
author = {L M A Bettencourt},
year = 2022,
publisher = {MIT Press},
address = {Cambridge MA}}

@book{von_neumann_automata,
title = {Theory of Self-Reproducing Automata},
author = {J Von Neumann and A. W. Burks},
year = 1966,
publisher = {University of Illinois Press},
address = {Urbana IL}}

@book{mathdestruction,
title = {Weapons of Math Destruction},
author = {C O'Neill},
year = 2016,
publisher = {Broadway Books},
address = {New York}}

@article{arrow1962economic,
 author = {Arrow, Kenneth J.},
 title = {The Economic Implications of Learning by Doing},
 journal = {Rev. Econ. Stud.},
 year = {1962},
 volume = {29},
 number = {3},
 pages = {155-173},
}

@book{hedstrom,
title = {Dissecting the Social},
author = {P Hedstr\"om},
year = 2005,
publisher = {Cambridge University Press},
address = {Cambridge}}

@book{giddens,
title = {The Constitution of Society},
author = {A Giddens},
year = 1984,
publisher = {Polity Press},
address = {Cambridge}}

@book{textbook_in_economic_game_theory,
title = {Game Theory},
author = {Michael Maschler and Shmuel Zamir and Eilon Solan},
year = 2013,
publisher = {Cambridge University Press},
address = {Cambridge}}

@book{friends,
 title = {Friends of Friends: Networks Manipulators and Coalitions},
 author = {J Boissevain},
 year = {1974},
 publisher = {Routledge},
address = {Oxford}
}

@book{simon_admin,
 title = {Administrative Behavior},
 author = {H A Simon},
 year = {1947},
 publisher = {Macmillan},
address = {New York}
}

@article{abbey1952,
 author = {Abbey, H.},
 title = {An examination of the {Reed-Frost} theory of epidemics},
 journal = {Hum. Biol.},
 year = {1952},
 volume = {24},
 pages = {201-233}
}

@book{axelrod1984,
 author = {Axelrod, R.},
 title = {The Evolution of Cooperation},
 year = {1984},
 publisher = {Basic Books},
 address = {New York, NY}
}

@article{watts2007twenty,
 author = {Watts, Duncan J.},
 title = {A twenty-first century science},
 journal = {Nature},
 year = 2007,
 volume = 445,
 pages = 489,
}

@article{hofman2021integrating,
 author = {Hofman, Jake M. and Watts, Duncan J. and Athey, Susan and Garip, Filiz and Griffiths, Thomas L. and Kleinberg, Jon and Margetts, Helen and Mullainathan, Sendhil and Salganik, Matthew J. and Vazire, Simine and Vespignani, Alessandro and Yarkoni, Tal},
 title = {Integrating explanation and prediction in computational social science},
 journal = {Nature},
 year = {2021},
 volume = {595},
 pages = {181-188},
}

@article{kalluri2025computer,
 author = {Kalluri, Pratyusha Ria and Agnew, William and Cheng, Myra and Owens, Kentrell and Soldaini, Luca and Birhane, Abeba},
 title = {Computer‑vision research powers surveillance technology},
 journal = {Nature},
 year = {2025},
pages = {73–79},
volume = 643,
}

@article{coleman1986,
 author = {Coleman, J. S.},
 title = {Social theory, social research, and a theory of action},
 journal = {Am. J. Sociol.},
 year = 1986,
 volume = 91,
 pages = {1309–1335}
}

@article{Carley01121994,
author = {Kathleen Carley and Allen Newell},
title = {The nature of the social agent},
journal = {J. Math. Sociol.},
volume = 19,
number = 4,
pages = {221-262},
year = 1994,
}

@incollection{frisch,
 author = {R Frisch},
 title = {Propagation problems and impulse problems in dynamic economics},
 booktitle = {Economic Essays in Honour of Gustav Cassel},
 editor = {K Kock},
 year = {1933},
 publisher = {Allen \& Unwin},
 address = {London},
 pages = {171–205}
}

@incollection{game_of_life,
 author = {P Rendell},
 title = {Turing universality of the Game of Life},
 booktitle = {Collision-Based Computing},
 editor = {A Adamatzky},
 year = {2002},
 publisher = {Springer},
 address = {London},
 pages = {513-541}
}

@incollection{plural_soar,
 author = {Kathleen Carley and Johan Kjaer-Hansen and Allen Newell and Michael Prietula},
 title = {PLURAL-{Soar}: A PROLEGOMENON TO ARTIFICIAL AGENTS AND ORGANIZATIONAL BEHAVIOR},
 booktitle = {Artificial Intelligence in Organization and Management Theory},
 editor = {Michae1 Masuch and Massimo Varglien},
 year = 1992,
 publisher = {North-Holland},
 address = {Amsterdam},
 pages = {87-118}
}

@book{newell_unified,
 author = {A Newell},
 title = {Unified Theories of Cognition},
 year = 1990,
 publisher = {Harvard University Press},
 address = {Cambridge MA}
}

@book{boehm,
 author = {Christopher Boehm},
 title = {Hierarchy in the Forest: The Evolution of Egalitarian Behavior},
 year = 1999,
 publisher = {Harvard University Press},
 address = {Cambridge MA}
}

@book{autopoiesis,
 author = {H Maturana and F Varela},
 title = {Autopoiesis and Cognition: The Realization of the Living},
 year = 1980,
 publisher = {D Reidel},
 address = {Dordrecht}
}

@book{pool1965,
 author = {De Sola Pool, I. and Abelson, R. P. and Popkin, S.},
 title = {Candidates, Issues, and Strategies: A Computer Simulation of the 1960 And 1964 Presidential Elections},
 year = 1965,
 publisher = {The MIT Press},
 address = {Cambridge MA}
}

@article{vc_ai,
 author = {Benedetta Montanaro and Annalisa Croce and Elisa Ughetto},
 title = {Venture capital investments in artificial intelligence},
 journal = {J. Evol. Econ.},
 year = 2024,
 volume = 34,
 pages = {1-28},
}

@book{thorndike1911animal,
 author = {Edward L. Thorndike},
 title = {Animal Intelligence},
 year = 1911,
 publisher = {Macmillan},
 address = {New York}
}

@article{charpentier2023reinforcement,
 author = {Charpentier, Arthur and \'{E}lie, Romuald and Remlinger, Carl},
 title = {Reinforcement Learning in Economics and Finance},
 journal = {Comput. Econ.},
 year = 2023,
 volume = 62,
 pages = {425-462},
}

@article{shirado_christakis,
 title = {Locally noisy autonomous agents improve global human coordination in network experiments},
 author = {Hirokazu Shirado and Nicholas A Christakis},
 journal = {Nature},
 volume = 545,
 pages = {370–374},
 year = 2020,
}

@article{silver2017mastering,
 title = {Mastering the game of go without human knowledge},
 author = {Silver, David and Schrittwieser, Julian and Simonyan, Karen and Antonoglou, Ioannis and Huang, Aja and Guez, Arthur and Hubert, Thomas and Baker, Lucas and Lai, Matthew and Bolton, Adrian and Yutian Chen and Timothy Lillicrap and Fan Hui and Laurent Sifre and George {van den Driessche} and Thore Graepel and Demis Hassabis},
 journal = {Nature},
 volume = 550,
 pages = {354-359},
 year = 2017,
}

@book{reinforcement,
 author = {Richard S. Sutton and Andrew G. Barto},
 title = {Reinforcement Learning: An Introduction},
 year = 2014,
edition = 2,
 publisher = {MIT Press},
 address = {Cambridge MA}
}

@article{dodds2003experimental,
 author = {Dodds, Peter Sheridan and Muhamad, Roby and Watts, Duncan J.},
 title = {An experimental study of search in global social networks},
 journal = {Science},
 year = 2003,
 volume = 301,
 number = 5634,
 pages = {827-829},
}

@book{dautenhahn,
 editor = {K Dautenhahn},
 title = {Human Cognition and Social Agent Technology},
 year = 2000,
 publisher = {John Benjamins},
 address = {Amsterdam}
}

@book{sugarscape,
 author = {Epstein, J. M. and Axtell, R.},
 title = {Growing Artificial Societies: Social Science from the Bottom Up},
 year = 1996,
 publisher = {Brookings Institution Press},
 address = {Washington DC}
}

@article{gilbert1966,
 author = {Gilbert, J. P. and Hammel, E. A.},
 title = {Computer simulation and analysis of problems in kinship and social structure},
 journal = {Am. Anthropol.},
 year = {1966},
 volume = {68},
 pages = {71-93}
}

@article{gullahorn1965,
 author = {Gullahorn, J. T. and Gullahorn, J. E.},
 title = {Some computer applications in social science},
 journal = {Am. Sociol. Rev.},
 year = {1965},
 volume = {30},
 pages = {353-365}
}

@article{lebaron2000agent,
 title = {Agent-based computational finance: Suggested readings and early research},
 author = {LeBaron, Blake},
 journal = {J. Econ. Dyn. Control},
 volume = {24},
 number = {5-7},
 pages = {679-702},
 year = {2000},
 publisher = {Elsevier}
}

@article{sawyer,
author = {R. Keith Sawyer},
title = {Artificial Societies: Multiagent Systems and the Micro-Macro Link in Sociological Theory},
journal = {Sociol. Method. Res.},
volume = {31},
number = {3},
pages = {325-363},
year = {2003},
}

@book{beltratti1996neural,
 title = {Neural Networks for Economic and Financial Modelling},
 author = {Beltratti, Andrea and Margarita, Sergio and Terna, Pietro},
 year = 1999,
address = {London},
 publisher = {International Thomson Computer Press}
}

@article{macy2002factors,
 author = {Macy, Michael W. and Willer, Robert},
 title = {From Factors to Actors: Computational Sociology and Agent-Based Modeling},
 journal = {Ann. Rev. Sociol.},
 year = 2002,
 volume = 28,
 pages = {143-166},
}

@article{lettau1997explaining,
 title = {Explaining the facts with adaptive agents: The case of mutual fund flows},
 author = {Lettau, Martin},
 journal = {J. Econ. Dyn. Control},
 volume = 21,
 number = 7,
 pages = {1117-1147},
 year = 1997,}

@book{gilbert_troitzsch,
 author = {N Gilbert and K G Troitzsch},
 title = {Simulation for the social scientist},
 year = 2005,
 publisher = {Open University Press},
 address = {New York}
}

@article{haigh2014alamos,
 title={Los {Alamos} bets on {ENIAC}: Nuclear {Monte Carlo} simulations, 1947-1948},
 author={Haigh, Thomas and Priestley, Mark and Rope, Crispin},
 journal={IEEE Ann. Hist. Comput.},
 volume=36,
 number=3,
 pages={42-63},
 year=2014,
}

@book{nilsson,
 author = {N J Nilsson},
 title = {The Quest for Artificial Intelligence: A History of Ideas and Achievements},
 year = 2009,
 publisher = {Cambridge University Press},
 address = {Cambridge}
}

@book{gilbert_conte,
 author = {N Gilbert and R Conte},
 title = {Artificial Societies: The Computer Simulation of Social Life},
 year = 1995,
 publisher = {Routledge},
 address = {London}
}

@book{hagerstrand1953,
 author = {H\"agerstrand, T.},
 title = {Innovationsförloppet ur Korologisk Synpunkt},
 year = 1953,
 publisher = {Gleerup},
 address = {Lund}
}

@article{desola1965,
 author = {{de Sola Pool}, I. and Kessler, A. R.},
 title = {The Kaiser, The Tsar, and The Computer: Information Processing in a Crisis},
 journal = {Am. Behav. Sci.},
 year = 1965,
 volume = 8,
 pages = {31-38}
}

@article{grimmer_roberts_stewart,
 author = "Grimmer, Justin and Roberts, Margaret E. and Stewart, Brandon M.",
 title = "Machine Learning for Social Science: An Agnostic Approach", 
 journal = "Annu. Rev. Polit. Sci.",
 year = 2021,
 volume = 24,
 pages = "395-419",
 }

@Incollection{Schatten2014,
author = "Schatten, Markus",
editor = "Magalh{\`a}es, Rodrigo",
title = "Structural Couplings of Organizational Design and Organizational Engineering",
bookTitle = "Organization Design and Engineering: Coexistence, Cooperation or Integration",
year = 2014,
publisher = "Palgrave Macmillan UK",
address = "London",
pages = "184-201",
}

@incollection{mead,
 author = {M Mead},
 title = {The Cybernetics of Cybernetics},
editor = {Heinz {von Foerster} and John D. White and Larry J. Peterson and John K. Russell},
 year = 1968,
booktitle = {Purposive Systems},
 pages = {1-11},
publisher = {Spartan Books},
address = {New York}
}

@article{early_email,
author = {James A. Danowski and Paul Edison-Swift},
title = {Crisis effects on intraorganizational computer-based communication},
journal = {Commun. Res.},
volume = 12,
number = 2,
pages = {251-270},
year = 1985,
}

@article{sagalnik_dodds_watts,
author = {Matthew J. Salganik and Peter Sheridan Dodds and Duncan J. Watts },
title = {Experimental Study of Inequality and Unpredictability in an Artificial Cultural Market},
journal = {Science},
volume = 311,
number = 5762,
pages = {854-856},
year = 2006,
}

@inproceedings{harman1961,
 author = {H H Harman},
 title = {Simulation: A review},
 year = 1961,
booktitle = {Proc.\ Western Joint IRE-AIEE-ACM Computer Conference},
 pages = {1-9}
}

@article{centaur,
 author = {Marcel Binz and Elif Akata and Matthias Bethge and Franziska Br\"andle and Fred Callaway and Julian Coda-Forno and Peter Dayan and Can Demircan and Maria K. Eckstein and Noémi \'{E}ltet\H{o} and Thomas L. Griffiths and Susanne Haridi and Akshay K. Jagadish and Li Ji-An and Alexander Kipnis and Sreejan Kumar and Tobias Ludwig and Marvin Mathony and Marcelo Mattar and Alireza Modirshanechi and Surabhi S. Nath and Joshua C. Peterson and Milena Rmus and Evan M. Russek and Tankred Saanum and Johannes A. Schubert and Luca M. Schulze Buschoff and Nishad Singhi and Xin Sui and Mirko Thalmann and Fabian J. Theis and Vuong Truong and Vishaal Udandarao and Konstantinos Voudouris and Robert Wilson and Kristin Witte and Shuchen Wu and Dirk U. Wulff and Huadong Xiong and Eric Schulz},
 title = {A foundation model to predict and capture human cognition},
 journal = {Nature},
volume = 644, 
pages = {1002–1009},
 year = 2025,
}

@article{hagendorff2023human,
 author = {Thilo Hagendorff and Sarah Fabi and Michal Kosinski},
 title = {Human-like intuitive behavior and reasoning biases emerged in large language models but disappeared in {ChatGPT}},
 journal = {Nat. Comput. Sci.},
 year = 2023,
 volume = 3,
 number = 10,
 pages = {833-838},
}

@article{granovetter1973strength,
 title = {The Strength of Weak Ties},
 author = {Granovetter, Mark S.},
 journal = {Am. J. Sociol.},
 volume = 78,
 number = 6,
 pages = {1360-1380},
 year = 1973,
}

@article{travers1969experimental,
 title = {An Experimental Study of the Small World Problem},
 author = {Travers, Jeffrey and Milgram, Stanley},
 journal = {Sociometry},
 volume = 32,
 number = 4,
 pages = {425-443},
 year = 1969,
}

@article{makovi2025rewards,
 author = {Makovi, Kinga and Bonnefon, Jean-Fran{\c c}ois and Oudah, Mayada and Sargsyan, Anahit and Rahwan, Talal},
 title = {Rewards and Punishments Help Humans Overcome Biases Against Cooperation Partners Assumed to be Machines},
 journal = {iScience},
 year = 2025,
 volume = 28,
 pages = 112833,
}

@article{kobis2025delegation,
  author       = {K{\"o}bis, Nils and Rahwan, Zeyad and Rilla, Randel and Supriyatno, Bimo I. and Bersch, Christian and Ajaj, Tarek and Bonnefon, Jean-Fran{\c{c}}ois and Rahwan, Iyad},
  title        = {Delegation to artificial intelligence can increase dishonest behaviour},
  journal      = {Nature},
  year         = 2025,
  volume       = 646,
  pages        = {126--134},
}

@article{argyle,
author = {Lisa P. Argyle and Ethan C. Busby and Joshua R. Gubler and Alex Lyman and Justin Olcott and Jackson Pond and David Wingate },
title = {Testing theories of political persuasion using {AI}},
journal = {Proc. Natl. Acad. Sci. USA},
volume = 122,
number = 18,
pages = {e2412815122},
year = 2025,
}

@article{science2025unethical,
 author = {Cathleen O'Grady},
 title = {`{U}nethical' {AI} research on {R}eddit under fire},
 journal = {Science},
 year = 2025,
volume = 388,
pages = {570-571}
}

@article{Hackenburg2025,
  author  = {Hackenburg, Kobi and Tappin, Ben M. and Hewitt, Luke and Saunders, Ed and Black, Sid and Lin, Hause and Fist, Catherine and Margetts, Helen and Rand, David G. and Summerfield, Christopher},
  title   = {The levers of political persuasion with conversational artificial intelligence},
  journal = {Science},
  year    = {2025},
  volume  = {390},
  number  = {6777},
  pages   = {eaea3884},
  doi     = {10.1126/science.aea3884}
}

@article{hackenburg2024evaluating,
 title = {Evaluating the persuasive influence of political microtargeting with large language models},
 author = {Hackenburg, Kobi and Margetts, Helen},
 journal = {Proc. Natl. Acad. Sci. USA},
 volume = 121,
 number = 24,
 pages = {e2403116121},
 year = 2024,
}

@article{tsvetkova_human_machine_networks,
author = {Tsvetkova, Milena and Yasseri, Taha and Meyer, Eric T. and Pickering, J. Brian and Engen, Vegard and Walland, Paul and L\"{u}ders, Marika and F\o{}lstad, Asbj\o{}rn and Bravos, George},
title = {Understanding Human-Machine Networks: A Cross-Disciplinary Survey},
year = 2017,
volume = 50,
number = 1,
journal = {ACM Comput. Surv.},
pages = 12,
}

@article{costello2024durably,
 author = {Thomas H. Costello and Gordon Pennycook and David G. Rand},
 title = {Durably reducing conspiracy beliefs through dialogues with {AI}},
 journal = {Science},
 year = 2024,
 volume = 385,
 number = 6714,
 pages = {eadq1814},
}

@unpublished{schmidhuber,
 author = {J Schmidhuber},
 title = {Annotated History of Modern {AI} and Deep Learning},
 note = {Preprint arXiv:2212.11279},
 year = 2022,
}

@unpublished{schroeder2025malicious,
 author = {Schroeder, Daniel Thilo and Cha, Meeyoung and Baronchelli, Andrea and Bostrom, Nick and Christakis, Nicholas A. and Garcia, David and Goldenberg, Amit and Kyrychenko, Yara and Leyton‐Brown, Kevin and Lutz, Nina and Marcus, Gary and Menczer, Filippo and Pennycook, Gordon and Rand, David G. and Schweitzer, Frank and Summerfield, Christopher and Tang, Audrey and Van Bavel, Jay and van der Linden, Sander and Song, Dawn and Kunst, Jonas R.},
 title = {How Malicious {AI} Swarms Can Threaten Democracy},
 note = {Preprint OSF/qm9yk\_v1},
 year = 2025,
}

@article{tsvetkova2024new,
 author = {Milena Tsvetkova and Taha Yasseri and Nicola Pescetelli and Tobias Werner},
 title = {A New Sociology of Humans and Machines},
 journal = {Nat. Hum. Behav.},
 year = 2024,
 volume = 8,
 pages = {1864-1876},
}

@article{mei2024turing,
 title = {A {Turing} test of whether {AI} chatbots are behaviorally similar to humans},
 author = {Mei, Qiaozhu and Xie, Yutong and Yuan, Walter and Jackson, Matthew O},
 journal = {Proc. Natl. Acad. Sci. USA},
 volume = 121,
 number = 9,
 pages = {e2313925121},
 year = 2024,
}

@article{froese2009,
 author = {Froese, Tom and Tom Ziemke},
 title = {Enactive Artificial Intelligence: Investigating the Systemic Organization of Life and Mind},
 journal = {Artif. Intell.},
 year = 2009,
 volume = 174,
number = {3-4},
 pages = {466-500}
}

@article{messeri2024artificial,
 author = {Messeri, Lisa and Crockett, Molly J.},
 title = {Artificial intelligence and illusions of understanding in scientific research},
 journal = {Nature},
 year = 2024,
 volume = 627,
 pages = {49-58},
}

@article{karjus,
 title = {Machine-assisted quantitizing designs: augmenting humanities and social sciences with artificial intelligence},
author = {A Karjus},
year = 2025,
journal = {Humanit. Soc. Sci. Commun.},
volume = 12,
pages = 277}

@inproceedings{focus_group,
author = {Zhang, Taiyu and Zhang, Xuesong and Cools, Robbe and Simeone, Adalberto},
title = {Focus Agent: {LLM}-Powered Virtual Focus Group},
year = 2024,
publisher = {Association for Computing Machinery},
address = {New York},
booktitle = {Proceedings of the 24th ACM International Conference on Intelligent Virtual Agents},
pages = 10,
location = {Glasgow, United Kingdom},
series = {IVA '24}
}

@article{farmer_abm,
 author = {Farmer, J Doyne},
 title = {Quantitative agent-based models: a promising alternative for macroeconomics},
 journal = {Oxf. Rev. Econ. Policy},
 pages = {graf027},
 year = 2025,
}

@article{gao2024,
 author = {Chen Gao and Xiaochong Lan and Nian Li and Yuan Yuan and Jingtao Ding and Zhilun Zhou and Fengli Xu and Yong Li},
 title = {Large language models empowered agent-based modeling and simulation: a survey and perspectives},
 journal = {Humanit. Soc. Sci. Commun.},
 year = 2024,
 volume = 11,
pages = 1259
}

@book{Boden2006,
 author = {Boden, Margaret A},
 title = {Mind as Machine: A History of Cognitive Science},
 publisher = {Oxford University Press},
 year = 2006,
 address = {Oxford},
}

@unpublished{wuttke2024,
 author = {Alexander Wuttke and Matthias Assenmacher and Christopher Klamm and Max M. Lang and Quirin W\"urschinger and Frauke Kreuter},
 title = {{AI} Conversational Interviewing: Transforming Surveys with {LLMs} as Adaptive Interviewers},
 note = {Preprint arXiv:2410.01824},
 year = 2024,
}

@unpublished{vaugrante2023,
 author = {Laurène Vaugrante and Mathias Niepert and Thilo Hagendorff},
 title = {A Looming Replication Crisis in Evaluating Behavior in Language Models? Evidence and Solutions},
 note = {Preprint arXiv:2305.03514},
 year = 2023,
}

@article{ziems2023can,
  author    = {Ziems, Caleb and Held, William and Shaikh, Omar and Chen, Jiaao and Zhang, Zhehao and Yang, Diyi},
  title     = {Can {Large Language Models} transform computational social science?},
  journal   = {Comput. Linguist.},
  volume    = {50},
  number    = {1},
  pages     = {237-291},
  year      = {2024},
}

@article{wilson,
 author = {M Wilson},
 title = {Six views of embodied cognition},
 journal = {Psychon. Bull. Rev.},
 year = 2002,
 volume = 9,
 pages = {625–636}
}

@book{simulacra,
 author = {J Baudrillard},
 title = {Simulacra and Simulation},
 year = 1994,
 publisher = {University of Michigan Press},
 address = {Ann Arbor MI}
}

@book{science_in_action,
 author = {B Latour},
 title = {Science in Action},
 year = 1987,
 publisher = {Harvard University Press},
 address = {Cambridge MA}
}

@book{ashby,
 author = {W R Ashby},
 title = {Design for a Brain},
 year = 1954,
 publisher = {John Wiley \& Sons},
 address = {New York}
}

@article{searle,
author = {J Searle},
title = {Minds, Brains, and Programs},
journal = {Behav. Brain Sci.},
year = 1980,
volume = 3,
number = 3,
pages = {417–457}
}

@book{hernan2020causal,
 author = {Miguel A. Hern\'an and James M. Robins},
 title = {Causal Inference: {W}hat If},
 year = 2024,
 edition = 2,
 publisher = {Chapman \& Hall / CRC},
 address = {Boca Raton},
}

@article{replication_crisis,
author = {{Open Science Collaboration}},
title = {Estimating the reproducibility of psychological science},
journal = {Science},
volume = 349,
number = 6251,
pages = {aac4716},
year = 2015,
}

@book{gibson,
 author = {J J Gibson},
 title = {The Senses Considered as Perceptual Systems},
 year = 1966,
 publisher = {Houghton Mifflin},
 address = {Boston MA}
}

@book{freeman2004development,
 title = {The Development of Social Network Analysis: A Study in the Sociology of Science},
 author = {Freeman, Linton C.},
 year = 2004,
 publisher = {Empirical Press},
 address = {Vancouver},
}

@book{minsky,
 author = {M Minsky},
 title = {The Society of Mind},
 year = 1986,
 publisher = {Simon \& Schuster},
 address = {New York}
}

@book{weizenbaum1976computer,
 author = {Joseph Weizenbaum},
 title = {Computer Power and Human Reason: From Judgment to Calculation},
 year = 1976,
 publisher = {W. H. Freeman},
 address = {San Francisco},
}

@book{meadows1972,
 author = {Meadows, D. H. and Meadows, D. L. and Randers, J. and Behrens, W. W.},
 title = {The Limits to Growth: A Report for the Club of Rome's Project on the Predicament of Mankind},
 year = 1972,
 publisher = {Universe Books},
 address = {New York},
}

@article{ward2010perils,
 title = {The perils of policy by p‐value: Predicting civil conflicts},
 author = {Ward, Michael D. and Greenhill, Brian D. and Bakke, Kristin M.},
 journal = {J. Peace Res.},
 volume = 47,
 number = 4,
 pages = {363-375},
 year = 2010,
}

@article{gautam2025survey,
 author = {Himanshu Gautam and Abhishek Gaur and Dharmendra Kumar Yadav},
 title = {A Survey on the Impact of Pre-Trained Language Models in Sentiment Classification Task},
 journal = {Int. J. Data Sci. Anal.},
 year = 2025,
}

@article{bala_goyal,
 title = {A NONCOOPERATIVE MODEL OF NETWORK FORMATION},
 author = {V Bala and S Goyal},
 journal = {Econometrica},
 volume = 68,
 number = 5,
 pages = {1181-1229},
 year = 2000,
}

@article{ghoshal_holme,
 title = {Dynamics of networking agents competing for high centrality and low degree},
 author = {P Holme and G Ghoshal},
 journal = {Phys. Rev. Lett.},
 volume = 96,
 number = 9,
 pages = 098701,
 year = 2006,
}

@incollection{luhmann1986autopoiesis,
 title = {The Autopoiesis of Social Systems},
 author = {Luhmann, Niklas},
 booktitle = {Sociocybernetic Paradoxes},
 editor = {Geyer, F. and {van der Zouwen}, J.},
 pages = {172-192},
 year = 1986,
 publisher = {SAGE Publications},
 address = {London}
}

@article{miller_revolution,
 author = {G Miller},
 title = {The cognitive revolution: a historical perspective},
 journal = {Trends Cogn. Sci.},
 year = 2003,
 volume = 7,
 pages = {141-144}
}

@article{newell1961,
 author = {Newell, A. and Simon, H. A.},
 title = {Computer simulation of human thinking},
 journal = {Science},
 year = 1961,
 volume = 134,
 pages = {2011-2017}
}

@article{simon_bounded,
 author = {H A Simon},
 title = {A Behavioral Model of Rational Choice},
 journal = {Q. J. Econ.},
 year = 1955,
 volume = 69,
number = 1,
 pages = {99-118}
}

@book{roberts,
 editor = {P C Roberts},
 title = {Modeling Large Systems},
 year = 1978,
 publisher = {Taylor \& Francis},
 address = {London}
}

@book{langton,
 editor = {C G Langton},
 title = {Artificial Life: Proceedings of an Interdisciplinary Workshop on the Synthesis and Simulation of Living Systems},
 year = 1989,
 publisher = {Routledge},
 address = {Milton Park}
}

@article{sayed,
year = 2014,
volume = 7,
journal = {Found. Trends Mach. Learn.},
title = {Adaptation, Learning, and Optimization over Networks},
number = {4-5},
pages = {311-801},
author = {Ali H. Sayed}
}

@book{arthur,
 author = {W Brian Arthur},
 title = {Complexity and the Economy},
 year = 2014,
 publisher = {Oxford University Press},
 address = {Oxford}
}

@book{aima,
 author = {Stuart J. Russell and Peter Norvig},
 title = {Artificial Intelligence: A Modern Approach},
 year = 2020,
 publisher = {Prentice Hall},
 address = {Hoboken NJ}
}

@article{turk1,
author = {John Bohannon},
title = {Social Science for Pennies},
journal = {Science},
volume = 334,
number = 6054,
pages = {307-307},
year = {2011},
}

@article{turk2,
author = {John Bohannon},
title = {Mechanical Turk upends social sciences},
journal = {Science},
volume = 352,
number = 6291,
pages = {1263-1264},
year = 2016,
}

@article{mturk_sux,
title = {The shape of and solutions to the {MTurk} quality crisis},
volume = 8,
number = 4,
journal = {Political Sci. Res. Methods.},
author = {Kennedy, Ryan and Clifford, Scott and Burleigh, Tyler and Waggoner, Philip D. and Jewell, Ryan and Winter, Nicholas J. G.},
year = 2020,
pages = {614–629}}

@article{pickard2011time,
 author = {G. Pickard and W. Pan and I. Rahwan and M. Cebrian and R. Crane and A. Madan and A. Pentland},
 title = {Time-critical social mobilization},
 journal = {Science},
 year = 2011,
 volume = 334,
 number = 6055,
 pages = {509-512},
}

@article{tang2011reflecting,
 author = {J. C. Tang and M. Cebrian and N. A. Giacobe and H. W. Kim and T. Kim and D. B. Wickert},
 title = {Reflecting on the {DARPA} red balloon challenge},
 journal = {Commun. ACM},
 year = 2011,
 volume = 54,
 number = 4,
 pages = {78-85},
}

@article{end_of_theory,
 title = {The end of theory: The data deluge makes the scientific method obsolete},
 author = {Anderson, Chris},
 journal = {Wired},
 volume = 16,
 number = 7,
 pages = {16-17},
 year = 2008
}

@book{simon1969,
 author = {Simon, H. A.},
 title = {The Sciences of the Artificial},
 year = 1969,
 publisher = {MIT Press},
 address = {Cambridge MA}
}

@book{simon_my_life,
 author = {Simon, H. A.},
 title = {Models of My Life},
 year = 1991,
 publisher = {Basic Books},
 address = {New York}
}

@article{anasazi,
author = {Robert L. Axtell and Joshua M. Epstein and Jeffrey S. Dean and George J. Gumerman and Alan C. Swedlund and Jason Harburger and Shubha Chakravarty and Ross Hammond and Jon Parker and Miles Parker},
title = {Population growth and collapse in a multiagent model of the {Kayenta Anasazi} in {Long House Valley}},
journal = {Proc. Natl. Acad. Sci. USA},
volume = 99,
pages = {7275-7279},
year = 2002}

@book{mitchell_artificial_2019,
	title = {Artificial {Intelligence}: {A} {Guide} for {Thinking} {Humans}},
	shorttitle = {Artificial {Intelligence}},
	publisher = {Penguin},
	author = {Mitchell, Melanie},
	year = 2019
}

@article{march_exploration_1991,
	title = {Exploration and exploitation in organizational learning},
	volume = 2,
	number = 1,
	journal = {Organ. Sci.},
	author = {March, James G.},
	year = 1991,
	pages = {71-87},
}

@book{schelling_micromotives_1978,
	title = {Micromotives and Macrobehavior},
	publisher = {W. W. Norton \& Company},
	author = {Schelling, Thomas C.},
	year = 1978
}

@article{march2021strategic,
 author = {March, Christina},
 title = {Strategic interactions between humans and artificial intelligence: Lessons from experiments with computer players},
 journal = {J. Econ. Psychol.},
 year = {2021},
 volume = {87},
 pages = {102426},
}

@inproceedings{engineering_prosociality,
author = {A Paiva and F P Santos and F C Santos},
title = {Engineering pro-sociality with autonomous agents},
year = {2018},
address = {Washington DC},
booktitle = {Proceedings of the Thirty-Second AAAI Conference
on Artificial Intelligence},
pages = {7994-7999},
}

@inproceedings{casa,
author = {Nass, Clifford and Steuer, Jonathan and Tauber, Ellen R.},
title = {Computers are social actors},
year = {1994},
address = {New York},
booktitle = {Proceedings of the SIGCHI Conference on Human Factors in Computing Systems},
pages = {72–78},
}

@article{sebo2020robots,
 author = {Sebo, Sarah and Stoll, Brian and Scassellati, Brian and Jung, Malte F.},
 title = {Robots in groups and teams: A literature review},
 journal = {Proc. ACM Hum.-Comput. Interact.},
 year = {2020},
 volume = {4},
 number = {CSCW2},
 pages = {176:1--176:36},
}

@article{burton2024how,
 author = {Burton, James W. and Lopez-Lopez, Eduardo and Hechtlinger, Simon and Rahwan, Zeyad and Aeschbach, Silvan and Bakker, Marten A. and Becker, Jan A. and Berditchevskaia, Anastasia and Berger, Jonah and Brinkmann, Lukas and Flek, Lucie and Herzog, Stefan M. and Huang, Shun and Kapoor, Shubham and Narayanan, Arvind and Nussberger, Ann-Marie and Yasseri, Taha and Nickl, Peter and Almaatouq, Abdullah and Hertwig, Ralph and others},
 title = {How large language models can reshape collective intelligence},
 journal = {Nat. Hum. Behav.},
 year = 2024,
 volume = 8,
 number = 9,
 pages = {1643-1655},
}

@article{vaccaro2024when,
 author = {Vaccaro, Matthew and Almaatouq, Abdullah and Malone, Thomas},
 title = {When combinations of humans and {AI} are useful: A systematic review and meta-analysis},
 journal = {Nat. Hum. Behav.},
 year = 2024,
 volume = 8,
 number = 12,
 pages = {2293-2303},
}

@article{tessler2024ai,
 author = {Tessler, Michael H. and Bakker, Marten A. and Jarrett, Daniel and Sheahan, Heather and Chadwick, Michael J. and Koster, Ralf and Evans, Greg and Campbell-Gillingham, Louise and Collins, Tom and Parkes, David C. and Botvinick, Matthew and Summerfield, Christopher},
 title = {{AI} can help humans find common ground in democratic deliberation},
 journal = {Science},
 year = 2024,
 volume = 386,
 number = 6719,
 pages = {eadq2852},
}

@article{noy2023experimental,
 author = {Noy, Shakked and Zhang, Whitney},
 title = {Experimental evidence on the productivity effects of generative artificial intelligence},
 journal = {Science},
 year = 2023,
 volume = 381,
 number = 6654,
 pages = {187-192},
}

@article{brynjolfsson2025generative,
 author = {Brynjolfsson, Erik and Li, Danielle and Raymond, Lindsey},
 title = {Generative {AI} at Work},
 journal = {Q. J. Econ.},
 year = 2025,
 volume = 140,
 number = 2,
 pages = {889-942},
}

@techreport{dellacqua2024navigating,
 author = {Dell'Acqua, Francesco and {McFowland III}, Edward and Mollick, Ethan R. and Lifshitz-Assaf, Hila and Kellogg, Katherine and Rajendran, Sandeep and Krayer, Laura and Candelon, François and Lakhani, Karim R.},
 title = {Navigating the Jagged Technological Frontier: Field Experimental Evidence of the Effects of AI on Knowledge Worker Productivity and Quality},
 institution = {Harvard Business School},
 year = 2024,
 number = {24-013},
}

@article{meincke2025chatgpt,
 author = {Meincke, Lasse and Nave, Gideon and Terwiesch, Christian},
 title = {ChatGPT decreases idea diversity in brainstorming},
 journal = {Nat. Hum. Behav.},
 year = 2025,
 volume = 9,
 number = 6,
 pages = {1107-1109},
}

@article{lee2024,
  author    = {Lee, B. C. and Chung, J.},
  title     = {An Empirical Investigation of the Impact of {ChatGPT} on Creativity},
  journal   = {Nat. Hum. Behav.},
  volume    = {8},
  pages     = {1906-1914},
  year      = 2024,
}

@article{farrell2025,
	title = {Large {AI} models are cultural and social technologies},
	volume = {387},
	doi = {10.1126/science.adt9819},
	number = {6739},
	journal = {Science},
	publisher = {American Association for the Advancement of Science},
	author = {Farrell, Henry and Gopnik, Alison and Shalizi, Cosma and Evans, James},
	month = mar,
	year = {2025},
	pages = {1153--1156},
}

@article{nguyen-trung2025,
	title = {{ChatGPT} in thematic analysis: {Can} {AI} become a research assistant in qualitative research?},
	volume = {59},
	issn = {1573-7845},
	shorttitle = {{ChatGPT} in thematic analysis},
	doi = {10.1007/s11135-025-02165-z},
	language = {en},
	number = {6},
	journal = {Quality \& Quantity},
	author = {Nguyen-Trung, Kien},
	month = dec,
	year = {2025},
	pages = {4945--4978},
}

@misc{horton2023,
	title = {Large language models as simulated economic agents: {What} can we learn from {Homo silicus}?},
	shorttitle = {Large {Language} {Models} as {Simulated} {Economic} {Agents}},
	url = {https://www.nber.org/papers/w31122},
	doi = {10.3386/w31122},
	urldate = {2026-07-03},
	publisher = {National Bureau of Economic Research},
	author = {Horton, John J. and Filippas, Apostolos and Manning, Benjamin S.},
	month = apr,
	year = {2023},
	doi = {10.3386/w31122}
}

@article{gottweis2026,
	title = {Accelerating scientific discovery with {Co}-{Scientist}},
	issn = {1476-4687},
	doi = {10.1038/s41586-026-10644-y},
	language = {en},
	urldate = {2026-07-03},
	journal = {Nature},
	publisher = {Nature Publishing Group},
	author = {Gottweis, Juraj and Weng, Wei-Hung and Daryin, Alexander and Tu, Tao and Sirkovic, Petar and Myaskovsky, Artiom and Glowaty, Grzegorz and Weissenberger, Felix and Orlandi, Alessio and Popovici, Dan and Palepu, Anil and Rong, Keran and Tanno, Ryutaro and Saab, Khaled and Zhang, Fan and Blum, Jacob and Carroll, Andrew and Kulkarni, Kavita and Tomašev, Nenad and Zverinski, Dina and Rendulic, Ivor and Vedadi, Elahe and Hasler, Florian and Rimanic, Luka and Boia, Marina and Budiselic, Ivan and Feinstein, Ben and Bellaiche, Mathias and Sheffer, Tom and Freyberg, Jan and Ratcliff, Jeremy and Bertolli, Ottavia and Chou, Katherine and Hassidim, Avinatan and Gokturk, Burak and Vahdat, Amin and Guan, Yuan and Dhillon, Vikram and Vaishnav, Eeshit Dhaval and Lee, Byron and Costa, Tiago R. D. and Penadés, José R. and Peltz, Gary and Matias, Yossi and Manyika, James and Hassabis, Demis and Xu, Yunhan and Kohli, Pushmeet and Pawlosky, Annalisa and Karthikesalingam, Alan and Natarajan, Vivek},
	month = may,
	year = {2026},
	pages = {1--10},
}

@article{Linetal2025,
  author  = {Lin, Hause and Czarnek, Gabriela and Lewis, Benjamin and White, Joshua P. and Berinsky, Adam J. and Costello, Thomas and Pennycook, Gordon and Rand, David G.},
  title   = {Persuading voters using human-artificial intelligence dialogues},
  journal = {Nature},
  year    = {2025},
  volume  = {648},
  number  = {8093},
  pages   = {394--401},
  doi     = {10.1038/s41586-025-09771-9}
}

\end{document}